\documentclass[conference]{IEEEtran}
\IEEEoverridecommandlockouts
\usepackage[utf8]{inputenc}
\usepackage{cite}
\usepackage{amsmath,amssymb,amsfonts}
\usepackage{algorithmic}
\usepackage{graphicx}
\usepackage{textcomp}
\usepackage{xcolor}
\usepackage{booktabs}
\usepackage{multirow}
\usepackage{subcaption}
\usepackage{makecell}
\usepackage{tabularx}
\usepackage{float}
\usepackage{url}

\newcommand{\nm}{{FlexST}}
\def\BibTeX{{\rm B\kern-.05em{\sc i\kern-.025em b}\kern-.08em
    T\kern-.1667em\lower.7ex\hbox{E}\kern-.125emX}}
\begin{document}

\title{A Multi-Resolution Multi-Domain Pre-Training Framework for Universal Traffic Forecasting

\thanks{$^*$ Work conducted during a visit to HKUST(GZ). $^\dag$ Corresponding authors.}
}

\author{\IEEEauthorblockN{Zhouyang Liu}
\IEEEauthorblockA{\textit{PDL Lab}\\
National University of Defense Technology\\
Hunan, China\\
liuzhouyang20@nudt.edu.cn}
\and
\IEEEauthorblockN{Jindong Han}
\IEEEauthorblockA{
\textit{Shandong University}\\
Shandong, China \\
jindong.han@sdu.edu.cn}
\and
\IEEEauthorblockN{Hao Wang}
\IEEEauthorblockA{
\textit{HKUST (GZ)}\\
hwang574@connect.hkust-gz.edu.cn}
\and
\IEEEauthorblockN{Xinyue Liu\\
Hui Gao}
\IEEEauthorblockA{
\textit{Didichuxing Co. Ltd}\\
Beijing, China \\
liuxinyue@didiglobal.com\\
deangaohui@didiglobal.com}
\and
\IEEEauthorblockN{Dongsheng Li$^\dag$}
\IEEEauthorblockA{\textit{PDL Lab} \\
\textit{National University of Defense Technology}\\
Hunan, China \\
dsli@nudt.edu.cn}
\and
\IEEEauthorblockN{Hao Liu$^\dag$}
\IEEEauthorblockA{
\textit{HKUST (GZ) and HKUST}\\
liuh@ust.hk}
}

\maketitle

\begin{abstract}
Spatio-temporal traffic data are central to intelligent transportation systems, yet their heterogeneity poses significant challenges for large-scale modeling. Existing pre-trained models often rely on a homogeneous modeling paradigm to handle highly heterogeneous traffic data. This fundamental mismatch not only limits model generalization but also leads to computationally expensive and parameter-inefficient designs. To this end, we propose \nm, a novel pre-training framework that introduces modularity and adaptivity for traffic modeling. Specifically, we first propose a multi-resolution spatio-temporal diffusion module that captures both short-term fluctuations and long-range trends, effectively reconciling inputs with divergent temporal and spatial resolutions. 
After that, we construct a domain-adaptive mixture-of-experts that dynamically routes data to specialized sub-networks, enabling selective knowledge transfer while preventing negative interference across diverse domains. Moreover, we devise a unified periodic encoding strategy that injects resolution- and domain-aware inductive biases to harmonize periodic inconsistencies across datasets. 
Extensive experiments on 23 real-world traffic datasets demonstrate that \nm\ significantly outperforms state-of-the-art baselines in zero- and few-shot settings, showcasing superior generalization, adaptability and efficiency. This work offers a new direction for building general-purpose pre-trained models capable of handling the complexity and variability of urban traffic systems. Code is available at \url{https://github.com/liuzhouyang/FlexST}.
\end{abstract}

\section{Introduction}

In recent years, the rapid deployment of urban sensing infrastructures (e.g., GPS-enabled mobile devices) has resulted in an explosion of traffic time series data. Accurate forecasting of such data is critical for optimizing real-time decision-making and improving urban mobility \cite{routeplanningref,congestion_moe}. While deep learning has achieved considerable success in modeling traffic time series, most existing models are trained on a single task or dataset~\cite{Song2020STSGCN,gwn,stid,staeformer}. Such specialized approaches struggle to generalize to diverse, data-scarce scenarios, leaving large amounts of traffic data underutilized and hindering the development of general-purpose intelligent transportation systems.

\begin{figure}[!ht]
    \centering
    \begin{subfigure}{\linewidth}
        \includegraphics[width=\linewidth]{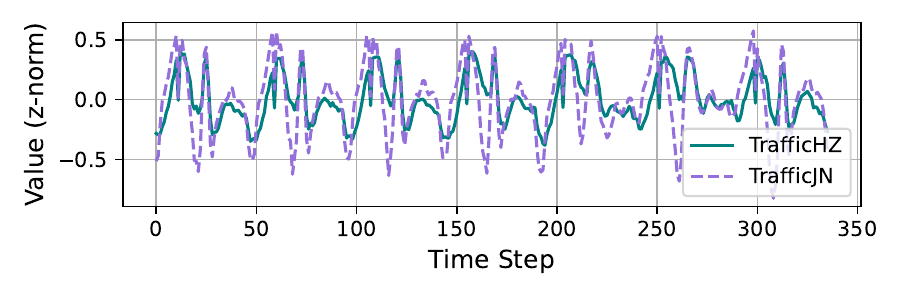}
    \end{subfigure}
    \vspace{-3mm} 
    \begin{subfigure}{\linewidth}
        \includegraphics[width=\linewidth]{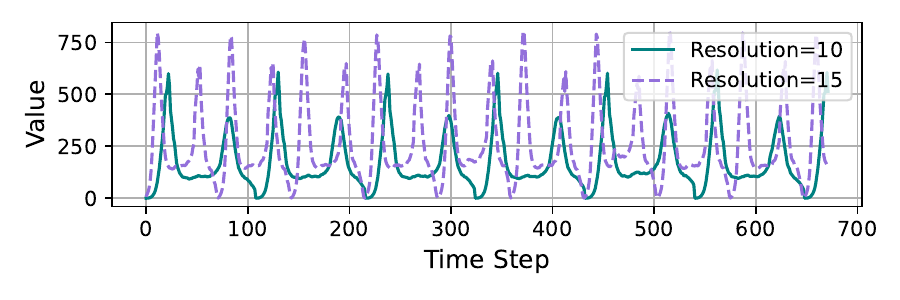}
    \end{subfigure}
    \vspace{-3mm}
    \begin{subfigure}{\linewidth}
        \includegraphics[width=\linewidth]{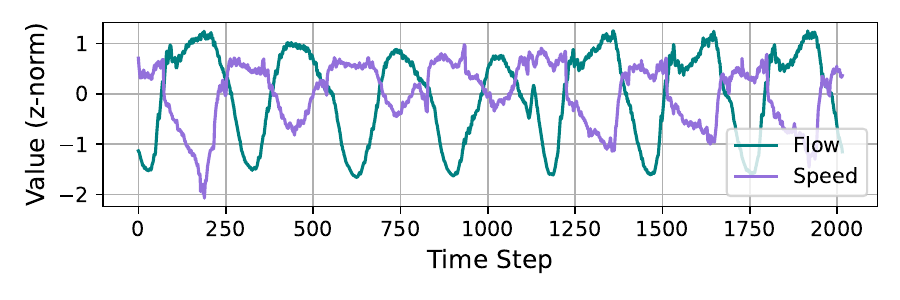}
    \end{subfigure}
    \caption{Weekly average traffic time series across regions, highlighting heterogeneity across cities (top: Hangzhou vs. Jinan), temporal resolutions (middle: Beijing Subway 10 min vs. 15 min), and variables (bottom: PEMS08 flow vs. speed). Z-score normalization handles large value differences.}\label{fig:hetero}
\end{figure}
Motivated by the success of Large Language Models (LLMs), researchers have started exploring the potential of large-scale pre-training for traffic time series. Early works~\cite{opencity,unist,uniflow,compactst,crossst} demonstrate that pre-training can capture shared traffic patterns across datasets, unlocking zero- and few-shot capabilities in previously unseen scenarios. However, existing studies typically adopt a homogeneous modeling paradigm, applying a single architecture and parameter space across all datasets. Unlike textual data, where syntactic and semantic regularities are relatively stable, traffic time series exhibit extreme heterogeneity across cities, resolutions, and variables (e.g., traffic speed, flow, and demand), as illustrated in Figure \ref{fig:hetero}. Treating them as homogeneous often results in negative knowledge transfer, which not only undermines predictive performance but also wastes computational resources by misallocating model capacity to conflicting patterns.

Specifically, this heterogeneity manifests in three fundamental challenges that hinder effective knowledge alignment: \textbf{(1) Spatio-temporal resolution mismatch}, where sampling intervals range from minutes to hours and spatial scales vary from individual road segments to region-level aggregations. Such discrepancies prevent a unified representation space, as identical observation windows may capture rapid local fluctuations in one dataset while reflecting only broad trends in another. \textbf{(2) Multi-domain heterogeneity}, where distinct traffic variables (e.g., speed, flow) and diverse urban contexts, including road networks, regulations, and mobility behaviors, create strong contextual differences. Forcing these into a shared parameter space often triggers negative transfer, where incompatible knowledge interferes with model learning and diminishes generalization. \textbf{(3) Cross-dataset periodic inconsistency}, where despite ubiquitous daily and weekly cycles \cite{largest,opencity,stid,staeformer}, variations in sampling and urban dynamics lead to phase shifts, amplitude variations, and inconsistent numbers of steps per period. Existing approaches that assume uniform periodicity \cite{opencity} or learn separate encodings \cite{informer,patchtst} fail to reconcile these diverse periodic behaviors across multi-resolution, multi-domain traffic data. 

To tackle the above challenges, we propose \nm, a modular and adaptive pre-training framework specifically designed to handle the heterogeneity of multi-resolution, multi-domain traffic time series data. \nm\ adopts a modular design that decouples dataset-specific variations while aligning transferable knowledge across datasets. It introduces three key components that strike a balance between commonality and diversity: (1) a multi-resolution spatio-temporal diffusion module that explicitly captures the varying traffic dynamics across datasets with different temporal and spatial resolutions; (2) a domain-adaptive mixture-of-experts that selectively activates submodels based on domain-specific characteristics, isolating incompatible patterns while retaining shared knowledge; and (3) a unified periodic encoding strategy that incorporates resolution- and domain-aware periodicity, addressing the inconsistencies in periodic patterns between datasets. Together, these components enable \nm\ to efficiently learn scalable representations, facilitate cost-effective knowledge transfer, and achieve robust generalization across a broad range of traffic scenarios. We summarize our contributions as follows.

\begin{itemize}
    \item We propose a multi-resolution spatio-temporal diffusion module, which explicitly captures traffic dynamics across diverse resolutions, enabling the model's adaptation to resolution-rich datasets.

    \item We introduce a domain-adaptive mixture-of-experts module, which selectively routes computations to isolate incompatible patterns while sharing reusable spatio-temporal knowledge across traffic datasets.

    \item We design a unified periodic encoding strategy, which injects resolution- and domain-aware periodic priors into the model, boosting its capability to capture cross-dataset traffic rhythms.

    \item We evaluate the propose framework on 23 real-world traffic datasets, demonstrating strong zero- and few-shot performance, along with extensive experiments validating the scalability and cost-effectiveness of our approach.
\end{itemize}

\section{Related Work}
\subsection{Traffic Forecasting}
Traffic forecasting is a longstanding spatio-temporal modeling problem. Evolution in traffic forecasting has moved from STGNNs~\cite{Song2020STSGCN, gwn} to advanced architectures like STID~\cite{stid} and STAEformer~\cite{staeformer}. 
Despite their effectiveness, these models are trained on a single dataset and exhibit limited flexibility in generalizing to diverse forecasting scenarios. To alleviate model bias and enhance robustness, Mixture-of-Experts (MoE) architectures have been explored ~\cite{congestion_moe,MH-MoE} . For instance, ST-MoE~\cite{stmoe} assigns experts to different road segment patterns. However, existing MoE-based methods remain task-specific and do not generalize across heterogeneous datasets. 
Motivated by the success of large-scale pre-training in natural language processing (NLP), recent studies~\cite{unist, opencity, uniflow,compactst,crossst} aim to develop transferable pre-trained spatio-temporal models for traffic forecasting. However, they often assume consistent spatio-temporal dependencies across multi-domain datasets, which rarely holds in practice. Consequently, these methods fall short of providing a flexible and adaptive modeling paradigm capable of accommodating diverse resolutions, variables, and urban contexts.
\begin{figure*}[!ht]
    \centering
    \includegraphics[width=0.95\linewidth]{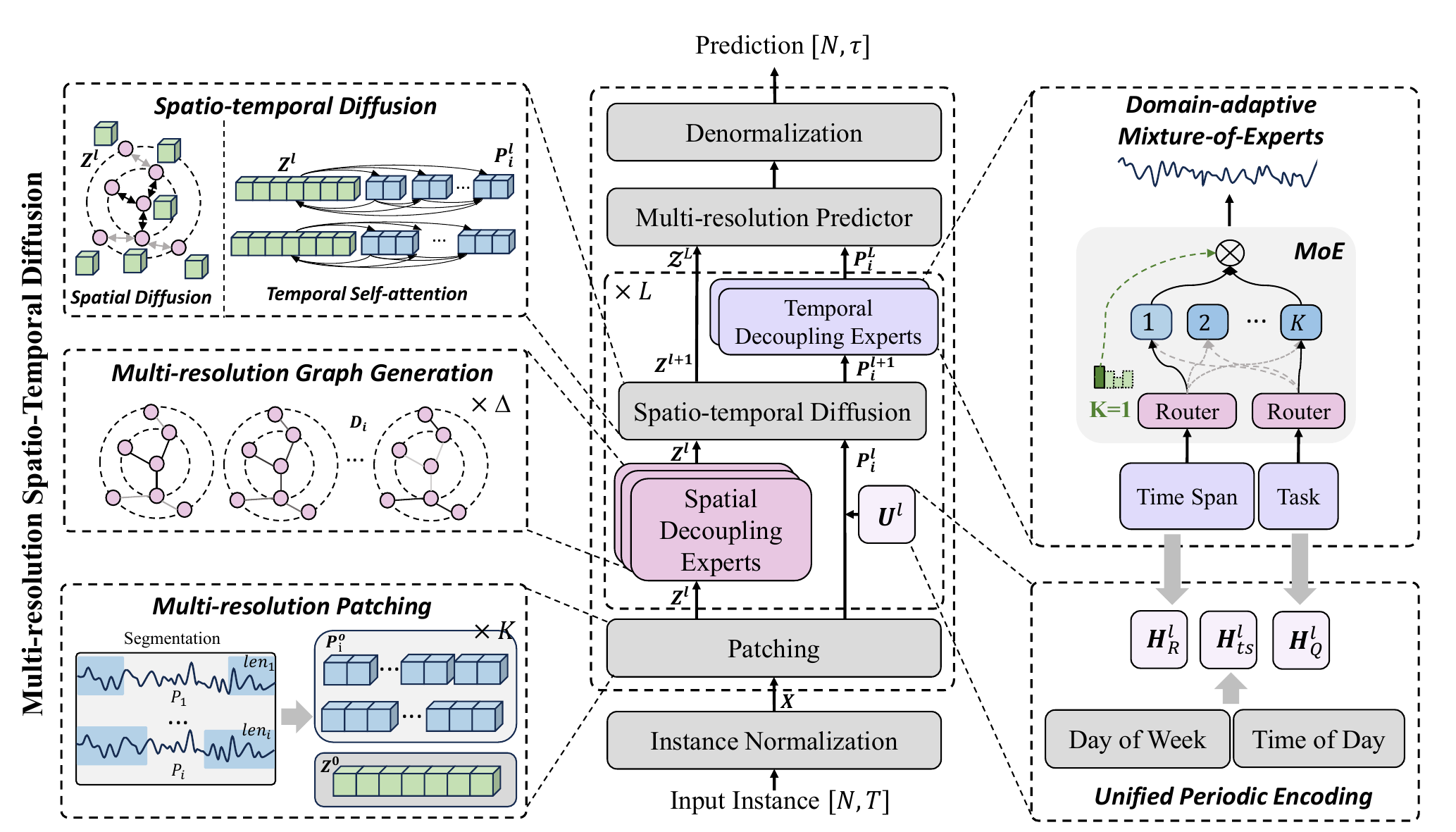}
    \caption{The model pipeline.}
    \label{fig:overview}
    \vspace{-5 mm}
\end{figure*}
\subsection{Time Series Forecasting}
Time series forecasting is critical across domains such as healthcare, finance, energy, and traffic. 
Earlier Transformer-based approaches operate on fixed-length input windows with self-attention to model long-range dependencies~\cite{informer,bigst}. Patching strategies segment input sequences into fixed or variable-length patches. For example, PatchTST~\cite{patchtst} leverages temporal patches to improve representation learning. MedFormer~\cite{medformer} further extends this idea by introducing multi-granularity temporal segmentation. 
Recently, a growing number of large-scale models pre-trained on extensive real-world or synthetic time series data have been developed~\cite{timer,timesfm,chronos,Moirai,UniTS,sundial}. Among them, Time-MoE \cite{timemoe}, incorporate mixture-of-experts mechanisms to dynamically route inputs based on data characteristics. However, these approaches primarily focus on temporal modeling alone, whereas traffic data involve complex spatio-temporal interactions, highlighting the need for frameworks that explicitly incorporate spatial structure during modeling and pre-training.

\subsection{Mixture-of-Experts (MoE)}
 MoE facilitates specialization and efficiency through dynamic input routing~\cite{moe0, moe1}, as evidenced by large-scale NLP models like GShard~\cite{gshard}. Beyond textual data, MoE has been adopted for multi-modal tasks~\cite{fusemoe}, which routes each modality to specialized experts, enabling effective multi-modal fusion while robustly handling missing inputs. These studies emphasize MoE's ability to reduce inference cost while learning specialized representations that flexibly adapt to heterogeneous or incomplete data. In contrast, our work introduces a multi-resolution, domain-adaptive MoE framework that aligns expert specialization with spatial and temporal dynamics across multi-domain datasets, enhancing transferability and generalization across heterogeneous forecasting settings.

\section{Preliminaries}
\paragraph{\textbf{Notion}} We represent traffic time series as a matrix $\mathbf{X}_{1:T} \in \mathbb{R}^{N \times T}$, where $N$ denotes the number of nodes (e.g., sensors, road segments, or regions) and $T$ is the observation period. Each row $\mathbf{x}_n = [x_{n,1}, \dots, x_{n,T}]$ corresponds to the time series of node $n$. We focus on graph-based traffic forecasting where the spatial topology is represented as $G = (\mathcal{V}, \mathbf{A})$, with $\mathcal{V}$ as the node set and $\mathbf{A} \in \mathbb{R}^{N \times N}$ as the adjacency matrix encoding pairwise spatial connectivity (e.g., geographical distances). 

\paragraph{\textbf{Single-domain traffic forecasting}} A single dataset is characterized by its task type $Q$, temporal resolution $R$ (sampling interval), spatial resolution $D$, and spatial graph $G$ (sensor, road link, or region). We define the spatial resolution $D$ via a propagation matrix $\mathbf{D}$, specifying the corresponding receptive field. Each observation in $\mathbf{X}_{1:T}$ is associated with timestamps $\mathbf{ts} \in \mathbb{R}^{N \times T \times 2}$ capturing time-of-day and day-of-week. Based on historical observations $\mathbf{X}_{1:T}$ with a parameterized model $f_{\theta}$, the goal of single-domain forecasting is to predict future conditions: 
\begin{equation}
    \mathbf{X}_{T+1:T+\tau} = f_{\theta}(\mathbf{X}_{1:T}).
\end{equation}
Here, $\{1:T\}$ is the look-back window and $\{T+1:T+\tau\}$ is the prediction horizon. The absolute time span $S$ of the prediction horizon is computed as $S = |\tau \cdot R|$, reflecting sequence duration across different resolutions.

\paragraph{\textbf{Multi-domain traffic data pre-training}}
We define multi-domain traffic data as a set of heterogeneous datasets $\mathcal{X} = \{\mathbf{X}_1, \dots, \mathbf{X}_J\}$. Each $\mathbf{X}_j$ is associated with its own spatial graph $G_j$, resolutions $(R_j, D_j)$, and objectives $Q_j$ (e.g., speed, flow, or demand). Our goal is to learn a general-purpose forecasting model $f_{\theta}$ from $\mathcal{X}$ that generalizes across diverse domains. We evaluate $f_{\theta}$ primarily on its zero-shot forecasting capability and its efficiency in few-shot adaptation through finetuning on unseen datasets.

\section{Methodology}
\subsection{Model Pipeline}
As illustrated in Figure~\ref{fig:overview}, \nm\ comprises three modules designed to reconcile traffic heterogeneity. 
(1) \emph{Multi-resolution Spatio-temporal Diffusion} serves as the backbone. It transforms raw series $\mathbf{X}$ and spatial graphs $G$ into multi-resolution patch representations $\mathbf{P}^0$, node-level global summaries $\mathbf{Z}^0$, and a set of heat diffusion-based propagation matrices $\mathcal{D}$. Stacked encoder layers then iteratively update $\mathbf{P}^l$ and $\mathbf{Z}^l$ through spatial diffusion and temporal self-attention to capture multi-resolution dynamics. 
(2) \emph{Domain-adaptive Mixture-of-Experts} enhances the backbone by dynamically routing representations to specialized sub-networks. Specifically, spatial experts select optimal propagation matrices $\mathbf{D}^l \in \mathcal{D}$ to update global summaries $\mathbf{Z}^l$, while temporal experts refine patch representations $\mathbf{P}^{l+1}$ after self-attention. This mechanism decouples domain-specific patterns and facilitates selective knowledge sharing. 
(3) \emph{Unified Periodic Encoding} augments input patches $\mathbf{P}^l$ at each layer by injecting resolution- and domain-aware periodic signals, resolving cross-dataset inconsistencies in periodicity. 

\subsection{Multi-resolution Spatio-temporal Diffusion}

This module adapts \nm\ to diverse resolutions by encoding both fine-grained fluctuations and long-term patterns. Following \cite{opencity}, we first apply instance normalization to each node independently to facilitate cross-context prediction. A denormalization step is applied post-prediction to restore the original scale.
\subsubsection{Multi-resolution patching}
To capture temporal dynamics across scales, \nm\ segments the normalized time series $\mathbf{X} \in \mathbb{R}^{N\times T}$ into patches of varying lengths. For each node $n \in \{1, \dots, N\}$, \nm\ generates a collection of patches $\mathbf{P}_i^\prime \in \mathbb{R}^{N \times M_i \times len_{i}}$ at each resolution level $i \in \{1, \dots, K\}$, where $len_{i}$ is the patch length and $M_i$ is the number of patches. To align these with the dataset's sampling interval $R$, we define resolution-specific time spans as $S_i = len_i\cdot R$. Each raw patch is then projected into a $d$-dimensional latent space and integrated with positional information:

\begin{align}
    \mathbf{P}^0_{i} = \text{Linear}(\mathbf{P}_i^\prime) + \mathbf{H}_{pos},
\end{align}
where $\mathbf{P}^0_{i} \in \mathbb{R}^{N\times M_i\times d}$ denotes the initial patch embeddings for resolution $i$, and $\mathbf{H}_{pos}$ represents the sinusoidal positional encoding~\cite{pos}. The resulting list of patch embedding sequences across all temporal resolutions is denoted as $\mathbf{P}^0$.

In parallel, \nm\ extracts a node-level global summary $\mathbf{Z}^0 \in \mathbb{R}^{N \times 1 \times d}$ by projecting the full input series through a linear layer: 
\begin{align}
    \mathbf{Z}^0 = \text{Linear}(\mathbf{X}).
\end{align}
While the patch embeddings $\mathbf{P}^0$ focus on localized fluctuations at multiple resolutions, the global summaries $\mathbf{Z}^0$ preserve the holistic context of the entire observation window. This joint representation ensures consistent coverage across datasets with heterogeneous sampling intervals.

\subsubsection{Multi-resolution graph generation}
To capture spatial multi-resolution, \nm\ generates multiple heat diffusion kernels, i.e., propagation matrices, inspired by~\cite{diffusion, grand, diffusion2}. Unlike standard graph convolutions, which are inherently limited to aggregating information from immediate $1$-hop neighbors, these matrices simulate a continuous-time diffusion process. This enables the model to achieve finer propagation across multiple hops in a single operation, bypassing the need for deep stacked layers that often trigger over-smoothing~\cite{diffusion, grand}. Given the adjacency matrix $\mathbf{A} \in \mathbb{R}^{N \times N}$ of graph $G$, we define a Laplacian-like operator:
\begin{align}
    \tilde{\mathbf{A}} = \mathbf{A} - \text{degree}(\mathbf{A}),
\end{align}
where $\text{degree}(\mathbf{A})$ is the degree matrix. By subtracting the self-contribution of each node, $\tilde{\mathbf{A}}$ emphasizes neighboring information propagation. Starting from $\mathbf{D}_{0} = \mathbf{I}$, \nm\ simulates diffusion at $\Delta$ predefined scales $\{\beta_i\}_{i=1}^{\Delta}\in [\beta_{\text{min}}, \beta_{\text{max}}]$, recursively computing the propagation matrix at resolution $i$ as:

\begin{equation}
    \mathbf{D}_{i} = e^{(\tilde{\mathbf{A}}\beta_i)} \mathbf{D}_{i-1}. \label{eq:kernel}
\end{equation}
Here, $e^{(\tilde{\mathbf{A}}\beta_i)}$ is the matrix exponential. By tuning the diffusion scale $\beta_i$, the model represents diverse spatial dynamics: smaller $\beta_i$ focuses on local propagation, while larger $\beta_i$ enables more global information flow. The resulting propagation matrix set $\mathcal{D} = \{\mathbf{D}_{i} \}_{i=1}^\Delta$ is precomputed to reduce computational cost, allowing \nm\ to adaptively learn spatial dynamics across diffusion resolutions.

\subsubsection{Spatio-temporal diffusion}
\nm\ employs an $L$-layer encoder to capture multi-resolution spatio-temporal dependencies. While spatio-temporal models typically propagate representations across the graph, directly diffusing all patch embeddings $\mathbf{P}$ is computationally expensive, as the cost grows proportionally with the number of patches. To optimize efficiency, \nm\ restricts spatial diffusion to the global summaries $\mathbf{Z}^l$. At each layer $l$, the global summaries are updated as:

\begin{align}
    \mathbf{Z}^l \leftarrow \text{MLP}(\mathbf{D}^l \mathbf{Z}^l) + \mathbf{Z}^l, \label{eq:spatial}
\end{align}
where $\text{MLP}$ is a multilayer perceptron. Here, $\mathbf{D}^l \in \mathcal{D}$ is a propagation matrix selected by the spatial experts. Specifically, these experts utilize a routing function to map the current global summaries $\mathbf{Z}^l$ and domain-specific metadata to the most effective diffusion resolution in $\mathcal{D}$, thereby reconciling spatial heterogeneity across datasets (see Section \ref{sec:spexpert}).

The spatially-enhanced $\mathbf{Z}^l$ is then appended to the patch representations $\mathbf{P}_i^l$ at each resolution $i$ and processed via temporal self-attention:
\begin{align}
    \mathbf{P}^{l+1}_i, \mathbf{Z}^{l+1}_i &= \text{SA}_i^{l}(\{\mathbf{P}^l_i+\mathbf{U}_i^l,\mathbf{Z}^{l}\}),\label{eq:sa}\\
    \mathbf{P}^{l+1}_i&\leftarrow\text{TDE}^l_i(\mathbf{P}^{l+1}_i),\label{eq:temporal}
\end{align}
where $\mathbf{U}_i^l$ is the periodic embedding from Section \ref{sec:upe}. $SA_i^l(\cdot)$ represents the temporal self-attention. The interaction in Eq. \eqref{eq:sa} enables mutual refinement: $\mathbf{P}_i^l$ absorbs spatial context, while $\mathbf{Z}^l$ is refined by temporal dependencies. The temporal experts $\text{TDE}^l_i(\cdot)$ then receive these outputs to decouple heterogeneous temporal dynamics. By routing inputs to specialized sub-networks based on their temporal characteristics, these experts isolate negative transfer and facilitate selective knowledge sharing, as detailed in Section \ref{sec:tmexpert}.

Finally, since global summaries $\{\mathbf{Z}^{l+1}_i\}_{i=1}^{K}$ capture resolution-specific dependencies, they are concatenated and projected into a unified representation:
\begin{align}
    \mathbf{Z}^{l+1} = \text{Linear}^l(\text{Concat}(\{\mathbf{Z}^{l+1}_i\}_{i=1}^{K})),
\end{align}
where $\mathbf{Z}^{l+1} \in \mathbb{R}^{N\times 1 \times d}$ integrates complementary information across spatio-temporal resolutions into a single global summary. The final output $\mathbf{P}^L$ and the unified summaries $\{\mathbf{Z}^l\}_{l=1}^L$ serve as the forecasting inputs.
\subsection{Domain-adaptive Mixture-of-Experts}

To address multi-domain heterogeneity, the domain-adaptive mixture-of-experts (MoE) operates on intermediate representations during spatial diffusion (Eq.~\ref{eq:spatial}) and temporal self-attention (Eq.~\ref{eq:temporal}). By introducing Temporal Decoupling Experts (TDE) and Spatial Decoupling Experts (SDE), \nm\ disentangles domain-specific variations while leveraging transferable knowledge. The sparse activation ensures that only a submodel is used per instance, reducing computational overhead and isolating incompatible patterns.

\subsubsection{Temporal decoupling experts} \label{sec:tmexpert}
TDE disentangles temporal dynamics through three expert types: (1) task-gated experts, (2) resolution-gated experts, and (3) a shared expert capturing universal patterns. These experts process patch representations in Eq. \ref{eq:temporal} to facilitate fine-grained temporal specialization. 

Each expert is a lightweight MLP. To promote scalable specialization, we adopt a sparse gating network where only the top-$K$ experts are activated. For a collection of experts $\mathcal{E}_I$ conditioned on $I$, the gating operates as:
\begin{align}
    \mathbf{s}_{I}^l &= \text{Softmax}(\mathbf{W}^l_{\mathcal{E}_I}I)\label{eq:score}, \\
    \mathbf{g}_{I}^l &= 
    \begin{cases} 
    \mathbf{s}_{I,i}^l, & \mathbf{s}_{I,i}^l \in \text{TopK}(\{\mathbf{s}_{I,j}^l \mid 0 \leq j < |\mathcal{E}_I|\}, K), \\ 
    0, & \text{otherwise}.
    \end{cases}\label{eq:topk}
\end{align}
Here, $\mathbf{s}{I}^l \in \mathbb{R}^{N \times M \times |\mathcal{E}I|}$ are scores for $M$ patches across $|\mathcal{E}_I|$ experts. For task-gated experts, $I$ is the task identity $Q$. For simplicity, we denote this gating process as $\text{gating}^l(I)$.

\begin{figure}[!ht]
\centering
\includegraphics[width=\linewidth]{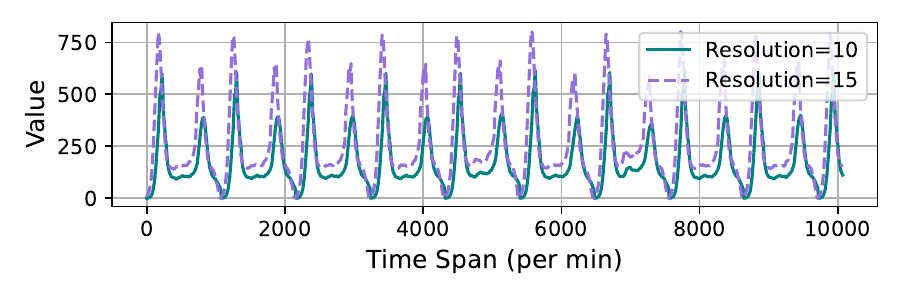}
\caption{Aligning different temporal resolutions by time span ensures that patches of the same absolute duration are compared consistently, regardless of their length or resolution.}\label{fig:align}
\end{figure}
For task-gated experts, $I$ is the task identity $Q$, i.e., a unique numeric ID. For resolution-gated experts, the time spans $\mathcal{S} = \{S_i\}_{i=1}^{K}$ serve as the condition. This allows patches with identical time spans from different resolutions to be aligned, as highlighted in Fig.~\ref{fig:align}, mitigating discrepancies from resolution mismatch. Separate gating networks compute relevance scores for task-gated $\mathcal{E}_Q$ and resolution-gated $\mathcal{E}_R$ expert groups:
\begin{align}
    \mathbf{g}_{Q}^l &= \text{gating}_Q^l(Q), \quad \mathbf{g}_{R}^l = \text{gating}_R^l(\mathcal{S}),\label{eq:specific_gated}
\end{align}
where $\mathbf{g}_{Q}^l \in \mathbb{R}^{N \times M \times |\mathcal{E}_Q|}$ and $\mathbf{g}_{R}^l \in \mathbb{R}^{N \times M \times |\mathcal{E}_R|}$ are gating scores.  This design balances specialized routing with robust generalization. For unseen tasks, task-gated experts provide fallback processing despite potentially less precise gating. Conversely, resolution-gated experts handle unseen resolutions by aligning patches via identical temporal spans, ensuring consistent dynamics across resolutions. Complementing these, the shared expert $E_{\text{share}}(\cdot)$ acts as a safety net, capturing universal, cross-domain patterns to ensure generalization to entirely unfamiliar tasks and resolutions.

The intermediate patch representations $\mathbf{P}^{l+1}$ are updated as a weighted sum of the outputs from selected task-gated experts, resolution-gated experts, and the shared expert:
\begin{align}
    \mathbf{P}^{l+1}_{Q} &=\sum_{i=1}^{|\mathcal{E}_Q|}\mathbf{g}_{Q,i}^l\cdot E_{Q,i}^l(\mathbf{P}^{l+1}),\\
    \mathbf{P}^{l+1}_{R} &=\sum_{i=1}^{|\mathcal{E}_R|}\mathbf{g}_{R,i}^l\cdot E_{R,i}^l(\mathbf{P}^{l+1}),\\
    \text{TDE}^l(\mathbf{P}^{l+1}) &= \alpha \cdot \mathbf{P}^{l+1}_{Q} + \beta \cdot \mathbf{P}^{l+1}_{R} + \gamma \cdot E_{share}(\mathbf{P}^{l+1}).
\end{align}
Here, learnable coefficients $\alpha, \beta, \gamma$ (summing to 1) control the relative contributions of each expert type. This process forms the temporally specialized embedding $\text{TDE}^l(\mathbf{P}^{l+1})$, leveraging decoupled, cross-domain transferable temporal patterns.

\subsubsection{Spatial decoupling experts}\label{sec:spexpert}
SDE disentangles the spatial dynamics via spatial experts. Each of them is paired with a distinct heat diffusion-based propagation matrix $\mathbf{D}_i \in \mathcal{D}$ computed in Eq.~\ref{eq:kernel}.  A spatial gate $\text{gating}^l(\mathbf{I})$, conditioned on task identity $Q$ and multi-resolution patch time span $\mathcal{S}$, selects the most relevant propagation matrix. To prevent the computational bottleneck of aggregating multiple spatial convolutions, \nm\ aggregates gate scores from task and resolution networks before applying top-1 selection: 
\begin{align}
    \mathbf{g}^l &= \text{gating}_Q^l(Q) + \text{gating}_R^l(\text{Concat}(\mathcal{S})),\\
     i^* &= \text{argmax}(\mathbf{g}^l),\\
    \mathbf{D}^l &= \mathbf{D}_{i^*}, \quad \mathbf{D}_{i^*} \in \mathcal{D},
\end{align}
where $i^*$ is the index of the highest-scoring matrix in $\mathcal{D}$. Unlike temporal experts, spatial experts are associated with matrices derived from the true adjacency matrix, ensuring they capture plausible propagation dynamics even for unseen tasks. 

\subsection{Unified Periodic Encoding}\label{sec:upe}
To resolve periodic inconsistencies across multi-domain datasets, \nm\ employs a unified additive encoding strategy that injects resolution- and domain-aware periodic biases into patch representations. This strategy integrates three types of context: task identity $Q$, patch time span $\mathcal{S}$, and timestamp information $\mathbf{ts}$ that encodes both the time-of-day (tod) and day-of-week (dow). \nm\ first maps $Q$ and $\mathcal{S}$ into $d$-dimensional latent spaces via embedding layers:
\begin{align}
     \mathbf{H}_Q^0 = \text{Emb}_Q(Q), \quad \mathbf{H}_{R}^0 = \text{Emb}_{R}(\mathcal{S}), 
\end{align}
where $\text{Emb}(\cdot)$ is a lookup table that maps discrete inputs to learnable embeddings. $\mathbf{H}_Q^0\in\mathbb{R}^{N \times d}$ and $\mathbf{H}_{R}^0\in\mathbb{R}^{N\times M \times d}$ are the initial task and resolution encodings. Notably, these embedding layers include extra placeholder slots to handle unseen tasks, resolutions, or missing timestamps, ensuring robustness in out-of-distribution scenarios or incomplete cases. 

To ensure consistency, each dataset shares the same timestamp encoding scheme: tod is indexed by the absolute time in minutes within a day, and dow by the weekday. Following OpenCity \cite{opencity}, \nm\ assigns the first frame of each patch as the representative timestamp $\mathbf{tod}_0, \mathbf{dow}_0\in\mathbb{R}^{N\times M}$, and computes initial timestamp encodings as follows:
\begin{align}
\mathbf{H}^0_{ts}=\text{Concat}(\text{Emb}_{tod}(\mathbf{tod}_0),\text{Emb}_{dow}(\mathbf{dow}_0) ),
\end{align}
where $\text{Emb}_{tod}(\cdot)$ and $\text{Emb}_{dow}(\cdot)$ embed time-of-day and day-of-week into $d/2$ dimensions, producing $\mathbf{H}^0_{ts} \in \mathbb{R}^{N \times M \times d}$ for multi-resolution patches, the timestamp encodings $\mathbf{H}^0_{ts,\tau} \in \mathbb{R}^{N \times 1 \times d}$ for prediction horizons are generated analogously. \nm\ then concatenates each patch's timestamp encoding with its targeted future:
\begin{align}
\mathbf{H}^0_{ts}\leftarrow\text{Concat}(\mathbf{H}^0_{ts},\mathbf{H}^0_{ts,\tau}),
\end{align}
Then, these contextual encodings are projected via separate linear layers and combined to form the unified periodic embedding:
\begin{align}
\mathbf{U}^l = \text{Linear}^l(\mathbf{H}_Q^l) + \text{Linear}^l(\mathbf{H}_R^l) + \text{Linear}^l(\mathbf{H}_{ts}^l).
\end{align}
Through this, task and resolution embeddings provide domain- and resolution-aware adjustments to the periodic information. Even for unseen tasks or temporal resolutions, the use of placeholders and absolute timestamp indexing allows \nm\ to leverage consistent periodic patterns by aligning unknown inputs with the shared periodic prior. The resulting unified representations $\mathbf{U}^l$ are added to patch representations in Eq. \ref{eq:sa}, serving as a domain-adaptive periodic prior across datasets.

\subsection{Multi-resolution Pre-training}\label{sec:pred} 
During pre-training, \nm, parameterized by $\theta$, takes as input a set of multi-domain traffic datasets $\mathcal{X} = \{\mathbf{X}_1, \dots, \mathbf{X}_J\}$ along with their associated spatial graph $G_j$, temporal resolution $R_j$, task type $Q_j$, and timestamps $\mathbf{ts}_j$. The encoder outputs the final multi-resolution patch representations $\mathbf{P}^L$ and the set of global summaries $\mathcal{Z}^L = \{\mathbf{Z}^{0}, \dots, \mathbf{Z}^{L}\}$ across all layers:
\begin{equation}
\mathbf{P}^L, \mathcal{Z}^L = \text{\nm}_{\theta}(\mathbf{X}_j, G_j, R_j, Q_j, \mathbf{ts}_j)
\end{equation}
For each resolution, a linear projection is applied to the collection $\{\mathbf{P}_i^L, \mathcal{Z}^L\}$ to obtain resolution-level predictions. These predictions are then concatenated and passed through a linear layer to produce the final prediction:
\begin{equation}
    \hat{X}_{T+1:T+\tau} = \text{Linear}\left(\text{Concat}\left(\{\text{Linear}_i(\{\mathbf{P}_i^L, \mathcal{Z}^L\})\}_{i=1}^{K}\right)\right).
\end{equation}
The parameters $\theta$ are optimized by minimizing the Mean Absolute Error (MAE) between predictions and ground-truth sequences:
\begin{equation}
\mathcal{L}_{\text{MAE}} = \frac{1}{N\cdot \tau}\sum_{n=1}^{N}\sum_{t=T+1}^{T+\tau}|X_{n:t} - \hat{X}_{n:t}|.
\end{equation}

\subsection{Complexity Analysis}
The model's total time complexity is dominated by $\mathcal{O}(N T d + L (N^2 d + \sum_i M_i^2 d + N M d^2))$, where $N$ is the number of nodes, $T$ the total time steps, $d$ the hidden dimension, $L$ the number of encoder layers, $M_i$ the number of patches at resolution $i$, and $M=\sum_i M_i$. The $\mathcal{O}(N^2 d)$ term from spatial diffusion can be reduced to near-linear using sparse or low-rank approximations, while $\sum_i M_i^2 d$ is manageable because $M_i \ll T$. Overall, the model remains efficient for large-scale multi-resolution spatio-temporal forecasting.
\section{Evaluations}
\subsection{Experimental Setup}
\subsubsection{Dataset} We conduct experiments on 23 traffic datasets covering traffic flow (PEMS03, PEMS04, PEMS07, PEMS08, SD, GBA), taxi demand (NYC-Taxi, Beijing Taxi, CHI-Taxi), bicycle sharing (NYC-Bike, CHI-Bike), traffic speed (TrafficHZ, TrafficJN, TrafficSH, TrafficZZ, PEMSBAY, METR-LA), metro flow (Beijing Subway 10 min, Beijing Subway 15 min, SHMetro, HZMetro), and traffic congestion level (DIDI-CD, DIDI-SZ). These datasets cover heterogeneous prediction tasks, spatial structures, and temporal resolutions. We pre-train models on 12 datasets and evaluate their zero-shot and few-shot capabilities on the remaining datasets. Detailed statistics are provided in supplementary materials.
\begin{table*}[ht]
\centering
\caption{Zero-shot performance on short-term prediction. We bold the \textbf{best} and underline the \underline{second} models. }\label{tab:zero_short}
\resizebox{\textwidth}{!}{ \renewcommand{\arraystretch}{1.0}
\setlength{\tabcolsep}{2pt}
\begin{tabular}{l|cc|cc|cc|cccc|cccc|cccc|cccc}
\toprule
\multirow{2}{*}{Dataset}  

& \multicolumn{2}{c|}{\multirow{2}{*}{\begin{tabular}[c]{@{}c@{}}PEMS04\end{tabular}}}

& \multicolumn{2}{c|}{\multirow{2}{*}{\begin{tabular}[c]{@{}c@{}}PEMS08\end{tabular}}}

& \multicolumn{2}{c|}{\multirow{2}{*}{\begin{tabular}[c]{@{}c@{}}Didi-SZ\end{tabular}}}

& \multicolumn{4}{c|}{Beijing Taxi}
& \multicolumn{4}{c|}{HZMetro}
& \multicolumn{4}{c|}{CHI-Bike}
& \multicolumn{4}{c}{Beijing Subway}\\
\cline{8-23}

& \multicolumn{2}{c|}{}  &
\multicolumn{2}{c|}{} &
\multicolumn{2}{c|}{} &
\multicolumn{2}{c}{Inflow} &
\multicolumn{2}{c|}{Outflow}&
\multicolumn{2}{c}{Inflow} &
\multicolumn{2}{c|}{Outflow}&
\multicolumn{2}{c}{Inflow} &
\multicolumn{2}{c|}{Outflow}&
\multicolumn{2}{c}{Inflow} &
\multicolumn{2}{c}{Outflow}\\
\midrule
&
\multicolumn{1}{c}{MAE} &
\multicolumn{1}{c|}{RMSE} &

\multicolumn{1}{c}{MAE} &
\multicolumn{1}{c|}{RMSE} &

\multicolumn{1}{c}{MAE} &
\multicolumn{1}{c|}{RMSE} &

\multicolumn{1}{c}{MAE} &
\multicolumn{1}{c}{RMSE} &

\multicolumn{1}{c}{MAE} &
\multicolumn{1}{c|}{RMSE} &

\multicolumn{1}{c}{MAE} &
\multicolumn{1}{c}{RMSE} &

\multicolumn{1}{c}{MAE} &
\multicolumn{1}{c|}{RMSE} &

\multicolumn{1}{c}{MAE} &
\multicolumn{1}{c}{RMSE} &

\multicolumn{1}{c}{MAE} &
\multicolumn{1}{c|}{RMSE} &

\multicolumn{1}{c}{MAE} &
\multicolumn{1}{c}{RMSE} &

\multicolumn{1}{c}{MAE} &
\multicolumn{1}{c}{RMSE} \\
\midrule
\midrule

\textbf{HA} &
38.67 &
57.49 &
 
31.93 &
47.72 &

3.19 &
4.81 &
 
56.39 &
94.83 &
 
56.47 &
94.93 &
 
130.30 &
235.99 &
 
130.00 &
219.45 &

6.02 &
11.18 &
5.99 &
11.53 &

238.66 & 
425.64 &
 
232.28 &
454.64 \\

\textbf{TimesFM} &
36.84 &
56.62 &
 
30.43 &
47.10 &

3.09 &
4.70 &
 
48.84 &
85.42 &
 
48.93 &
85.53 &

113.21 &
216.89 &
 
114.88 &
204.20 &
 
5.69 &
10.57 &
 
5.63 &
10.83 &
 
201.74 &
385.23 &
 
197.52 &
415.69 \\

\textbf{Timer} &
125.88 &
152.52 &
 
117.77 &
141.59 &

7.83 &
10.58 &
 
90.80 &
127.08 &
 
90.84 &
127.18 &
 
150.13 &
258.28 &

150.24 &
245.23 &
 
6.24 &
12.91 &
 
6.24 &
13.23 &
 
224.25 &
425.30 &
 
219.84 &
463.39 \\

\textbf{Sundial} &
36.56 &
55.35 &
 
30.20 &
46.14 &

3.18 &
4.84 &

50.90 &
88.25 &
 
51.01 &
88.38 &
 
123.18 &
221.96 &

124.59 &
221.96 &
 
5.85 &
10.99 &
 
5.82 &
11.35 &
 
224.25 &
425.30 &
 
219.84 &
463.39 \\

\textbf{Time-MoE} &
42.54 &
62.40 &
 
34.78 &
51.06 &

3.79 &
5.68 &
 
64.01 &
107.72 &
 
64.11 &
107.85 &
 
150.48 &
278.39 &

148.70 &
256.23 &
 
6.92 &
12.88 &
 
6.85 &
13.24 &
 
276.56 &
502.13 &
 
271.56 &
544.85 \\

\textbf{OpenCity$_{mini}$} &
26.85 &
43.03 &
 
21.46 &
34.74 &

2.87 &
4.44 &

45.59 &
86.09 &
 
45.69 &
86.26 &
 
87.65 &
180.65 &
 
85.67 &
165.69 &

5.37 &
10.81 &
5.37 &
11.42 &
 
135.17 & 
294.76 &
 
131.41 &
314.01 \\

\textbf{OpenCity$_{base}$} &
26.80 &
42.87 &
 
21.49 &
34.87 &

2.88 &
4.45 &
 
\underline{44.17} &
\underline{81.87} &
 
\underline{44.33} &
\underline{82.11} &
 
89.89 &
188.08 &
 
91.54 &
174.36 &

5.39 &
10.70 &
5.37 &
11.17 &

147.34 & 
310.94 &
 
141.59 &
326.84 \\

\textbf{OpenCity$_{plus}$} &
26.76 &
42.92 &
 
21.45 &
34.98 &

2.92 &
4.51 &
 
47.05 &
93.10 &
 
47.18 &
93.26 &
 
90.75 &
190.28 &
 
87.12 &
168.13 &

5.62 &
11.47 &
5.65 &
12.23 &

134.84 & 
302.42 &
 
132.81 &
321.31 \\

\textbf{CrossST} &
32.82 &
52.71 &
 
29.11 & 
41.91 &
 
4.03 & 
5.92 &
 
52.41 & 
88.73 &
 
52.47 & 
88.83 &

90.81 & 
\textbf{176.38} &
 
92.61 & 
169.62 &

\textbf{4.88} &
\underline{10.03} &
\textbf{4.74} &
\underline{10.31} &

\underline{114.30} & 
\textbf{237.28} &
 
\underline{119.50} &
\underline{291.88} \\

\textbf{CompactST} &
\underline{26.46} & 
\underline{42.21} &
 
\underline{21.07} & 
\textbf{34.12} &
 
\underline{2.83} & 
\underline{4.39} &
 
48.65 & 
91.57 &
 
48.73 & 
91.69 &
 
\underline{85.24} & 
\underline{177.46} &
 
\underline{83.48} & 
\textbf{161.66} &

5.65 &
11.44 &
5.60 &
11.93 &

126.71 & 
281.97 &
 
125.28 &
307.33 \\
\midrule
\textbf{Ours} &
\textbf{24.68} &
\textbf{40.51} &
 
\textbf{20.00} &
\underline{34.61} &
 
\textbf{2.63} &
\textbf{4.10} &
 
\textbf{36.32} &
\textbf{67.72} &
 
\textbf{36.46} &
\textbf{67.96} &
 
\textbf{78.23} &
180.03 &
 
\textbf{73.69} &
\underline{166.55} &

\underline{4.89} &
\textbf{8.99} &
\underline{4.88} &
\textbf{9.37} &

\textbf{107.94} & 
\underline{237.34} &
 
\textbf{103.50} &
\textbf{265.58} \\
\textbf{Improv. (\%)} &
6.73 &
4.03 &
 
5.08 &
-1.44 &
 
7.07 &
6.61 &
 
17.77 &
17.28 &
 
17.75 &
17.23 &
 
8.22 &
-2.07 &
 
11.73 &
-3.02 &
 
0.20 &
10.37 &
 
-2.95 &
9.12 &
 
5.56 &
-0.03 &
 
13.39 &
9.01 \\
\bottomrule                                            
\end{tabular}}
\label{tab:res}
\end{table*}

\begin{table}[ht]
\centering
\caption{Zero-shot performance on long-term prediction.}\label{tab:zero_long}
\resizebox{\linewidth}{!}{ \renewcommand{\arraystretch}{1.0}
\setlength{\tabcolsep}{2pt}
\begin{tabular}{l|cc|cc|cc|cccc}
\toprule
\multirow{2}{*}{Dataset}  

& \multicolumn{2}{c|}{\multirow{2}{*}{\begin{tabular}[c]{@{}c@{}}PEMS04\end{tabular}}}

& \multicolumn{2}{c|}{\multirow{2}{*}{\begin{tabular}[c]{@{}c@{}}PEMS08\end{tabular}}}

& \multicolumn{2}{c|}{\multirow{2}{*}{\begin{tabular}[c]{@{}c@{}}Didi-SZ\end{tabular}}}
& \multicolumn{4}{c}{HZMetro}\\
\cline{8-11}

& \multicolumn{2}{c|}{}  &
\multicolumn{2}{c|}{} &
\multicolumn{2}{c|}{} &
\multicolumn{2}{c}{Inflow} &
\multicolumn{2}{c}{Outflow}\\
\midrule
&
\multicolumn{1}{c}{MAE} &
\multicolumn{1}{c|}{RMSE} &

\multicolumn{1}{c}{MAE} &
\multicolumn{1}{c|}{RMSE} &

\multicolumn{1}{c}{MAE} &
\multicolumn{1}{c|}{RMSE} &

\multicolumn{1}{c}{MAE} &
\multicolumn{1}{c}{RMSE} &

\multicolumn{1}{c}{MAE} &
\multicolumn{1}{c}{RMSE} \\
\midrule
\midrule

\textbf{HA} &
123.73 &
160.18 &
 
104.67 &
137.75 &

5.04 &
6.86 &

106.77 &
193.77 &
 
106.63 &
183.58 
\\

\textbf{TimesFM} &
95.63 &
132.00 &
 
72.03 &
105.05 &

4.28 &
5.96 &

87.759 &
218.089 &
 
78.832 &
182.676 \\

\textbf{Timer} &
111.52 &
132.88 &
 
106.06 &
128.96 &
 
6.39 &
8.82 &

142.714 &
242.061 &
 
145.531 &
236.142 \\

\textbf{Sundial} &
101.46 &
136.81 &
 
67.46 &
97.97 &

4.82 &
6.50 &
 
159.44 &
313.21 &
 
157.96 &
304.78 \\

\textbf{Time-MoE} &
137.62 &
177.34 &
 
116.32 &
152.45 &
 
6.45 &
8.75 &

140.09 & 
266.19 &
 
141.14 &
245.64 \\

\textbf{OpenCity$_{mini}$} &
\underline{50.39} &
\underline{78.57} &
 
\underline{39.43} &
\underline{63.46} &

3.44 &
5.13 &

47.87 &
113.24 &
 
48.84 &
116.64 \\

\textbf{OpenCity$_{base}$} &
51.26 &
80.89 &
 
40.18 &
65.07 &

3.29 &
4.95 &

47.21 &
\underline{107.52} &
 
46.08 &
\textbf{103.38} \\

\textbf{OpenCity$_{plus}$} &
53.01 &
82.67 &
 
40.78 &
64.85 &

\underline{3.26} &
\underline{4.90} &

\underline{46.13} &
112.89 &
 
\underline{46.05} &
114.41 \\

\textbf{CrossST} &
62.23 &
93.44 &
 
58.49 &
80.39 &
 
6.97 &
9.65 &
 
79.73 &
143.77 &
 
79.98 &
134.08 \\
 
 \textbf{CompactST} &
56.18 &
87.15 &
 
45.08 &
71.83 &
 
3.97 &
5.86 &
 
55.10 &
133.32 &
 
52.96 &
118.30 \\
\midrule
\textbf{Ours} &
\textbf{47.52} &
\textbf{74.13} &
 
\textbf{36.45} &
\textbf{57.26} &

\textbf{3.11} &
\textbf{4.74} &

\textbf{42.52} &
\textbf{100.17} &
 
\textbf{43.44} &
\underline{104.05} \\

\bottomrule                                            
\end{tabular}}
\label{tab:res}
\end{table}

\subsubsection{Baselines} We evaluate zero-shot forecasting against representative pre-trained spatio-temporal models and time-series foundation models, including OpenCity$_{mini}$, OpenCity$_{base}$, OpenCity$_{plus}$ \cite{opencity}, CrossST \cite{crossst}, CompactST \cite{compactst}, TimesFM \cite{timesfm}, Timer \cite{timer}, Sundial \cite{sundial} and Time-MoE \cite{timemoe}. UniST \cite{unist} and UniFlow \cite{uniflow} are excluded because UniST is restricted to grid-based data and UniFlow requires graph partitioning, which would hinder fair evaluation. For few-shot evaluation, we include strong classical and deep learning baselines, including Historical Average (HA), Informer \cite{informer}, PatchTST \cite{patchtst}, STGCN \cite{Song2020STSGCN}, GWNET \cite{gwn}, ST-Norm \cite{stnorm}, STID \cite{stid}, and STAEformer \cite{staeformer}. We adopt official implementations and hyper-parameter settings when available, and otherwise use the BasicTS library \cite{basicst} for consistent and reproducible evaluation.

\subsubsection{Implementation details} For short-term prediction, we use multi-resolution patching with $K=1$ and a patch length of 12. For long-term prediction, we set $K=2$ with patch lengths of 8 and 64. For all resolutions, the stride equals the patch length, resulting in non-overlapping temporal patches. In multi-resolution graph generation, we construct 13 propagation matrices by linearly sampling diffusion scales $\beta_i \in [0.1, 3.0]$. For domain-adaptive mixture-of-experts, top-$1$ experts are selected for both spatial and temporal modules, with a total of 9 temporal decoupling experts. Each expert is implemented as a MLP with two linear layers and a ReLU activation, using dimensions $(64, 32, 64)$. The hidden dimension $d$ is set to 64, the encoder depth is 3, post-norm with RMSNorm is applied, and the dropout rate is fixed at 0.1. Hyperparameter choices are detailed in the supplementary materials.

\begin{table*}[ht]
\centering
\caption{Few-shot performance on short-term prediction using training data of 7 days.}\label{tab:few12}
\resizebox{\textwidth}{!}{ \renewcommand{\arraystretch}{1.0}
\setlength{\tabcolsep}{2pt}
\begin{tabular}{l|cc|cc|cc|cccc|cccc|cccc|cccc}
\toprule
\multirow{2}{*}{Dataset}  

& \multicolumn{2}{c|}{\multirow{2}{*}{\begin{tabular}[c]{@{}c@{}}PEMS08\end{tabular}}}

& \multicolumn{2}{c|}{\multirow{2}{*}{\begin{tabular}[c]{@{}c@{}}Didi-SZ\end{tabular}}}

& \multicolumn{2}{c|}{\multirow{2}{*}{\begin{tabular}[c]{@{}c@{}}TrafficZZ\end{tabular}}}

& \multicolumn{4}{c|}{NYC-Taxi}
& \multicolumn{4}{c|}{CHI-Taxi}
& \multicolumn{4}{c|}{NYC-Bike}
& \multicolumn{4}{c}{CHI-Bike}\\
\cline{8-23}

& \multicolumn{2}{c|}{}  &
\multicolumn{2}{c|}{} &
\multicolumn{2}{c|}{} &
\multicolumn{2}{c}{Inflow} &
\multicolumn{2}{c|}{Outflow}&
\multicolumn{2}{c}{Inflow} &
\multicolumn{2}{c|}{Outflow}&
\multicolumn{2}{c}{Inflow} &
\multicolumn{2}{c|}{Outflow}&
\multicolumn{2}{c}{Inflow} &
\multicolumn{2}{c}{Outflow}\\
\midrule
&
\multicolumn{1}{c}{MAE} &
\multicolumn{1}{c|}{RMSE} &

\multicolumn{1}{c}{MAE} &
\multicolumn{1}{c|}{RMSE} &

\multicolumn{1}{c}{MAE} &
\multicolumn{1}{c|}{RMSE} &

\multicolumn{1}{c}{MAE} &
\multicolumn{1}{c}{RMSE} &

\multicolumn{1}{c}{MAE} &
\multicolumn{1}{c|}{RMSE} &

\multicolumn{1}{c}{MAE} &
\multicolumn{1}{c}{RMSE} &

\multicolumn{1}{c}{MAE} &
\multicolumn{1}{c|}{RMSE} &

\multicolumn{1}{c}{MAE} &
\multicolumn{1}{c}{RMSE} &

\multicolumn{1}{c}{MAE} &
\multicolumn{1}{c|}{RMSE} &

\multicolumn{1}{c}{MAE} &
\multicolumn{1}{c}{RMSE} &

\multicolumn{1}{c}{MAE} &
\multicolumn{1}{c}{RMSE} \\
\midrule
\midrule

\textbf{HA} &
32.19 &
48.07 & 
 
3.17 &
4.79 & 
 
1.46 &
2.09 &
 
16.99 &
38.73 & 
 
25.15 &
51.74 &
 
4.42 &
11.53 & 
 
4.80 &
12.46 &
 
14.47 &
10.48 & 

14.30 &
24.16 &
 
6.16 &
11.54 & 
 
5.99 &
11.54 \\

\textbf{Informer} &
69.10 &
94.23 &
 
3.05 &
4.70 & 
 
1.09 &
1.79 &
 
18.54 &
41.33 & 
 
30.07 &
57.14 &
 
4.13 &
13.15 & 
 
4.62 &
14.81 &
 
13.45 &
24.82 & 

13.42 &
24.50 &
 
4.37 &
8.89 & 
 
4.27 &
8.76 \\

\textbf{PatchTST} &
22.26 &
35.58 &
 
2.82 &
4.34 & 
 
1.16 &
1.80 &
 
11.37 &
26.91 & 
 
17.25 &
38.02 &
 
3.30 &
8.42 & 
 
3.60 &
9.40 &
 
10.87 &
19.77 & 

10.86 &
19.34 &
 
5.00 &
9.68 & 
 
4.88 &
9.55 \\

\textbf{STGCN} &
28.99 &
42.40 &
 
3.39 &
5.09 & 
 
1.07 &
1.69 &
 
17.71 &
38.17 & 
 
28.68 &
52.78 &
 
3.66 &
11.17 & 
 
4.10 &
13.04 &
 
11.11 &
21.87 & 

11.10 &
21.66 &
 
\underline{3.53} &
\underline{7.24} & 
 
\underline{3.52} &
\underline{7.18} \\

\textbf{GWNET} &
\underline{19.97} &
\underline{31.61} &
 
2.74 &
4.19 & 
 
1.07 &
1.61 &
 
12.43 &
28.64 & 
 
22.32 &
43.36 &
 
\underline{3.19} &
10.34 & 
 
\underline{3.53} &
11.38 &
 
10.50 &
21.47 & 

10.46 &
21.09 &
 
\textbf{3.50} &
7.32 & 
 
3.58 &
7.55 \\

\textbf{STNorm} &
24.84 &
38.42 &
 
\underline{2.64} &
\underline{4.05} & 
 
0.94 &
1.46 &
 
15.11 &
33.56 & 
 
24.44 &
45.88 &
 
3.76 &
12.17 & 
 
4.28 &
13.80 &
 
12.72 &
24.46 & 

12.63 &
24.06 &
 
3.81 &
7.86 & 
 
3.75 &
7.86 \\

\textbf{STID} &
20.71 &
33.13 &
 
2.92 &
4.45 & 
 
1.74 &
2.31 &
 
11.67 &
27.27 & 
 
17.52 &
37.17 &
 
3.89 &
11.20 & 
 
4.22 &
12.25 &
 
10.50 &
20.27 & 

10.47 &
19.89 &
 
5.06 &
10.00 & 
 
4.96 &
10.02 \\

\textbf{STAEformer} &
42.21 &
69.05 &
 
2.75 &
4.21 & 
 
\underline{0.93} &
\underline{1.42} &
 
16.17 &
38.97 & 
 
27.65 &
54.94 &
 
3.71 &
12.22 & 
 
4.13 &
13.60 &
 
11.23 &
22.29 & 

11.37 &
22.23 &
 
3.79 &
7.81 & 
 
3.86 &
8.10 \\
\textbf{Ours (Zero)} &
20.05 &
34.74 &
 
\underline{2.64} &
4.12 & 
 
1.09 &
1.68 &
 
\underline{8.62} &
\underline{20.30} & 
 
\underline{12.70} &
\underline{27.53} &
 
3.60 &
\underline{8.34} & 
 
3.77 &
\underline{8.94} &
 
\underline{10.48} &
\underline{19.32} & 

\underline{10.57} &
\underline{19.27} &
 
5.01 &
9.32 & 
 
4.88 &
9.37 \\
\midrule
\textbf{Ours (Few)} &
\textbf{18.23} & 
\textbf{30.36} & 
 
\textbf{2.58} & 
\textbf{4.03} &
 
\textbf{0.89} &
\textbf{1.38} & 
 
\textbf{8.34} & 
\textbf{19.54} & 
 
\textbf{12.46} &
\textbf{27.00} & 
 
\textbf{2.90} & 
\textbf{7.08} &
 
\textbf{3.25} &
\textbf{8.26} & 
 
\textbf{7.66} & 
\textbf{13.84} & 

\textbf{7.43} &
\textbf{13.30} &
 
\underline{3.53} & 
\textbf{6.75} & 
 
\textbf{3.48} & 
\textbf{6.56} \\ 

\bottomrule                                            
\end{tabular}}
\end{table*}

\begin{table}[ht]
\centering
\caption{Few-shot performance on long-term prediction using training data of 7 days.}\label{tab:few64}
\resizebox{\linewidth}{!}{ \renewcommand{\arraystretch}{1.0}
\setlength{\tabcolsep}{2pt}
\begin{tabular}{l|cc|cc|cc|cccc|cccc}
\toprule
\multirow{2}{*}{Dataset}  

& \multicolumn{2}{c|}{\multirow{2}{*}{\begin{tabular}[c]{@{}c@{}}PEMS08\end{tabular}}}

& \multicolumn{2}{c|}{\multirow{2}{*}{\begin{tabular}[c]{@{}c@{}}Didi-SZ\end{tabular}}}

& \multicolumn{2}{c|}{\multirow{2}{*}{\begin{tabular}[c]{@{}c@{}}TrafficZZ\end{tabular}}}
& \multicolumn{4}{c|}{NYC-Taxi}
& \multicolumn{4}{c}{NYC-Bike}\\
\cline{8-15}

& \multicolumn{2}{c|}{}  &
\multicolumn{2}{c|}{} &
\multicolumn{2}{c|}{} &
\multicolumn{2}{c}{Inflow} &
\multicolumn{2}{c|}{Outflow}&
\multicolumn{2}{c}{Inflow} &
\multicolumn{2}{c}{Outflow}\\
\midrule
&
\multicolumn{1}{c}{MAE} &
\multicolumn{1}{c|}{RMSE} &

\multicolumn{1}{c}{MAE} &
\multicolumn{1}{c|}{RMSE} &

\multicolumn{1}{c}{MAE} &
\multicolumn{1}{c|}{RMSE} &

\multicolumn{1}{c}{MAE} &
\multicolumn{1}{c}{RMSE} &

\multicolumn{1}{c}{MAE} &
\multicolumn{1}{c|}{RMSE} &

\multicolumn{1}{c}{MAE} &
\multicolumn{1}{c}{RMSE} &

\multicolumn{1}{c}{MAE} &
\multicolumn{1}{c}{RMSE} \\
\midrule
\midrule

\textbf{HA} &
105.80 &
139.29 & 
 
4.95 &
6.75 & 
 
1.18 &
1.66 &
 
13.87 &
31.39 & 
 
20.99 &
42.47 &
 
11.86 &
20.41 & 

11.66 &
20.04 \\

\textbf{Informer} &
87.83 &
118.34 &
 
3.69 &
5.38 & 
 
1.09 &
1.77 &
 
19.56 &
43.31 & 
 
31.40 &
59.84 &
 
13.45 &
24.50 & 
 
13.37 &
24.13 \\

\textbf{PatchTST} &
46.88 &
73.00 &
 
3.35 &
5.07 & 
 
\textbf{0.59} &
\textbf{0.95} &
 
7.31 &
18.71 & 
 
11.45 &
27.58 &
 
7.44 &
14.28 & 
 
7.68 &
14.55 \\

\textbf{STGCN} &
51.81 &
74.70 &
 
4.11 &
5.82 & 
 
1.09 &
1.72 &
 
22.84 &
45.49 & 
 
34.75 &
60.09 &
 
10.73 &
21.24 & 
 
10.66 &
20.81 \\

\textbf{GWNET} &
40.54 &
59.04 &
 
3.31 &
5.04 & 
 
1.09 &
1.74 &
 
13.60 &
31.47 & 
 
20.59 &
42.28 &
 
11.74 &
21.72 & 
 
11.61 &
21.30 \\

\textbf{STNorm} &
66.54 &
89.31 &
 
4.13 &
5.91 & 
 
1.19 &
1.80 &
 
23.26 &
45.39 & 
 
34.59 &
59.06 &
 
13.85 &
25.32 & 
 
13.77 &
24.94 \\

\textbf{STID} &
40.26 &
62.36 &
 
3.96 &
5.57 & 
 
2.57 &
3.38 &
 
11.20 &
21.54 & 
 
15.78 &
29.98 &
 
10.30 &
17.54 & 
 
10.19 &
17.29 \\

\textbf{STAEformer} &
47.47 &
70.55 &
 
\underline{3.00} &
\underline{4.52} & 
 
\underline{0.63} &
\underline{0.98} &
 
11.96 &
27.60 & 
 
18.64 &
38.96 &
 
9.50 &
19.63 & 
 
9.49 &
19.39 \\
\textbf{Ours (Zero)} &
\underline{36.62} &
\underline{57.61} &
 
3.10 &
4.74 & 
 
0.67 &
1.06 &
 
\underline{5.03} &
\underline{11.97} & 
 
\underline{7.66} &
\underline{17.11} &
 
\underline{6.60} &
\underline{12.40} & 

\underline{6.70} &
\underline{12.64} \\
 
\midrule
\textbf{Ours (Few)} &
\textbf{29.30} & 
\textbf{49.56} & 
 
\textbf{2.89} & 
\textbf{4.43} & 
 
\textbf{0.59} & 
\textbf{0.95} &
 
\textbf{5.02} & 
\textbf{11.96} & 
 
\textbf{7.65} & 
\textbf{17.08} & 
 
\textbf{6.55} & 
\textbf{12.33} & 

\textbf{6.67} &
\textbf{12.57} \\
 
\bottomrule                                            
\end{tabular}}
\end{table}

\subsubsection{Evaluation Metrics} Following the previous work \cite{unist,opencity}, we adopt \emph{Root Mean Square Error (RMSE)} and \emph{Mean Absolute Error (MAE)} to measure the prediction errors. Similar to Mean Square Error (MSE), RMSE is sensitive to outliers but presents errors in the original units for better interpretability, while MAE measures the average magnitude of errors.

\subsubsection{Experimental Protocol} All models are implemented in PyTorch and trained on an NVIDIA A40 GPU. For pre-training, datasets are split into training, validation, and test sets with a ratio of 3:2:2. For few-shot prediction, seven days of training data are used, with 20\% reserved for validation and the remainder for testing. Following \cite{unist}, both the historical and prediction horizons are set to 12 for short-term and 64 for long-term prediction. The batch size is 64. Pre-training is conducted for up to 100 epochs, while few-shot training and finetuning are limited to 30 epochs. We use the AdamW optimizer with a learning rate of 0.001, apply gradient clipping with a threshold of 5, and adopt early stopping with a patience of 15 based on validation loss. Due to high pre‑training cost, we run experiments only once, focusing on cross‑dataset consistency as in standard large‑scale pre‑training.
\begin{figure}[t]
    \centering
    \includegraphics[width=\linewidth]{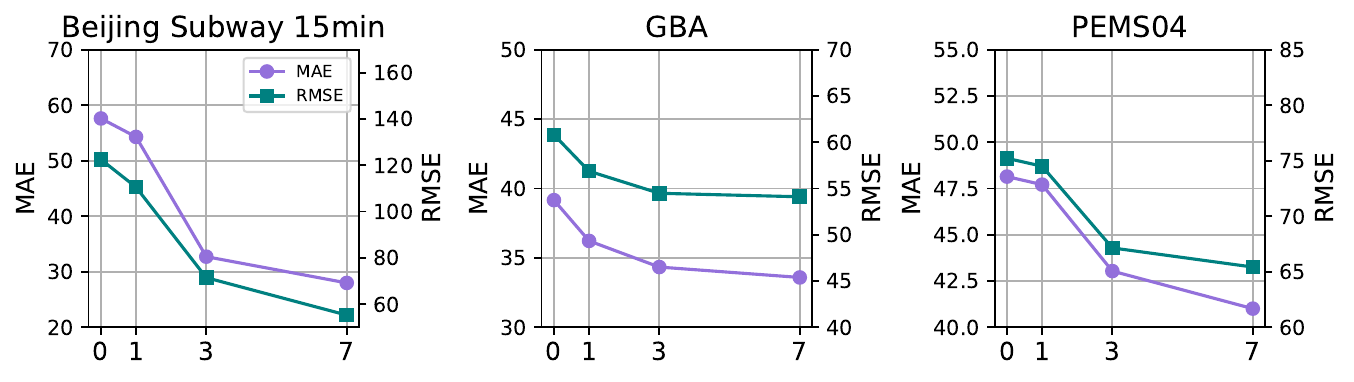}
    \caption{Comparative performance evaluation of few-shot prediction with varying training data volumes (1, 3, and 7 days).}
    \label{fig:ft}
    \vspace{-5mm}
\end{figure}
\begin{figure*}[t]
    \centering
    \begin{subfigure}[b]{0.32\textwidth}
        \centering
        \includegraphics[width=\textwidth]{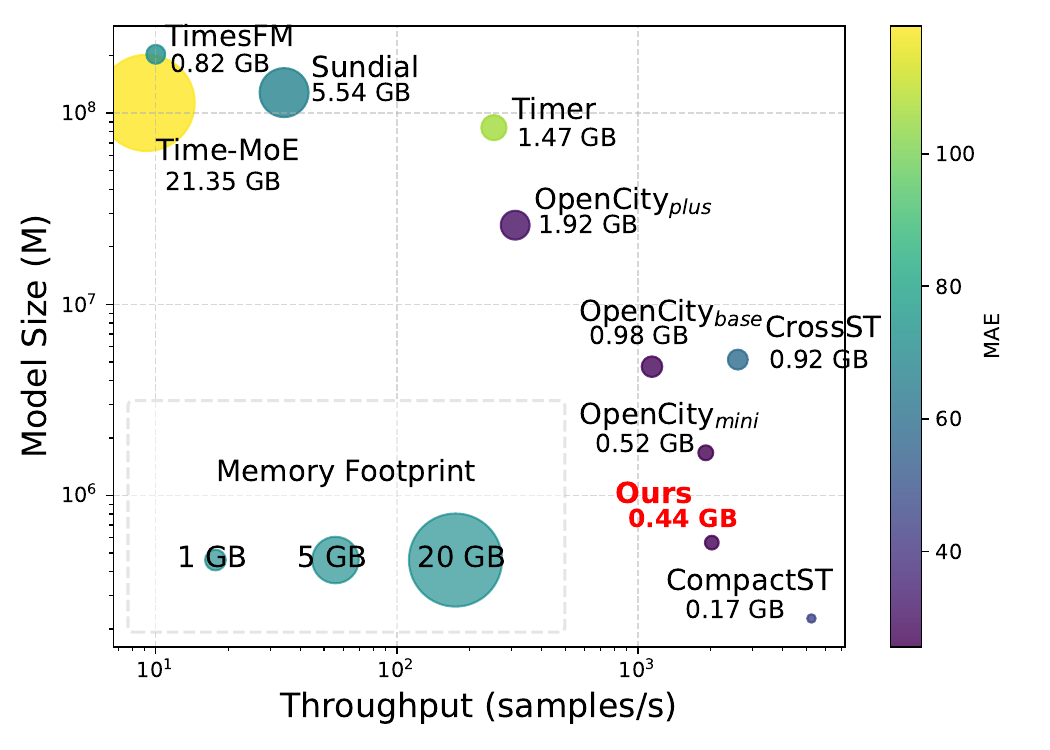}\\
        \includegraphics[width=\textwidth]{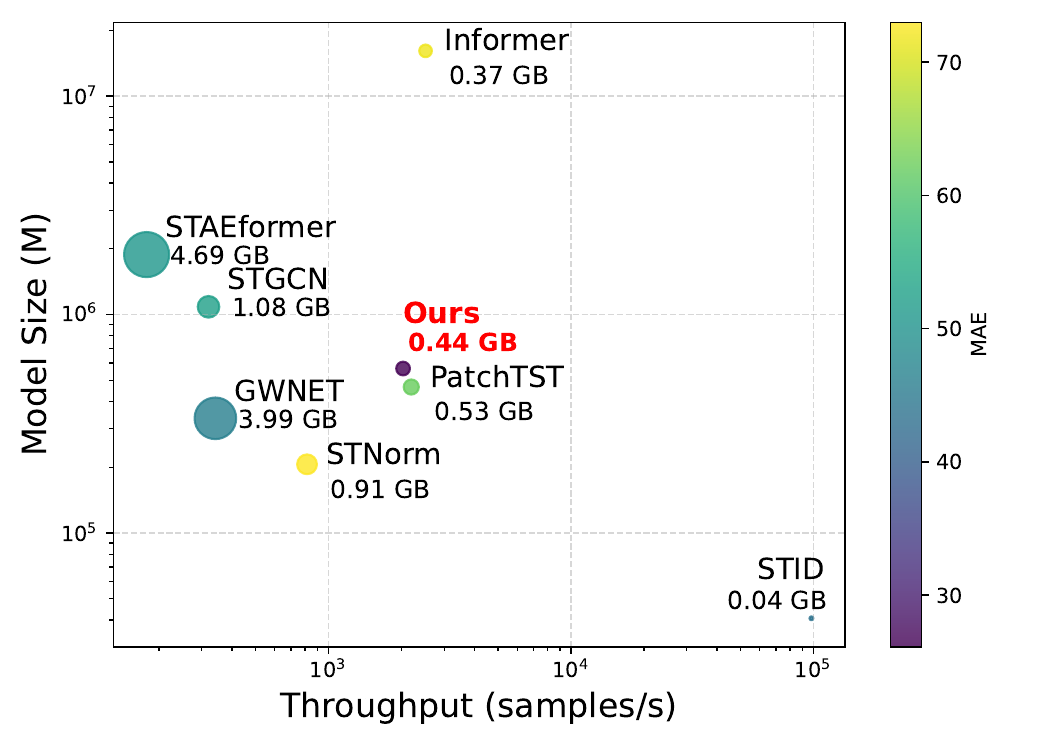}
        \caption{PEMS08}\label{fig:sub1}
    \end{subfigure}
    \begin{subfigure}[b]{0.32\textwidth}
        \centering
        \includegraphics[width=\textwidth]{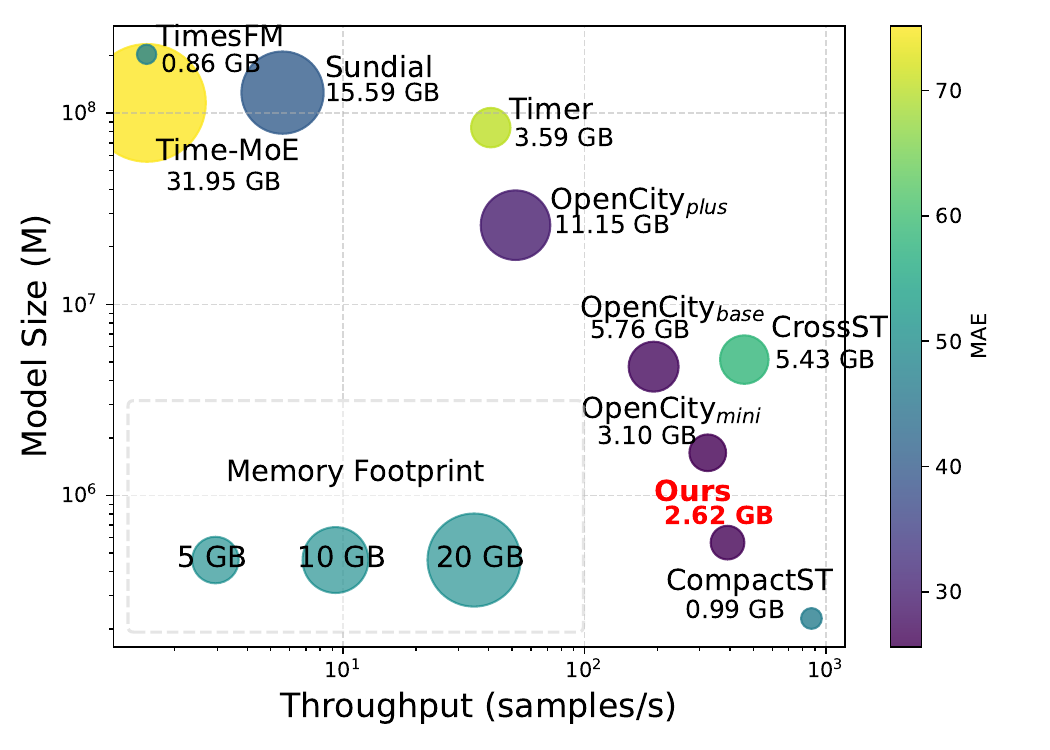}\\
        \includegraphics[width=\textwidth]{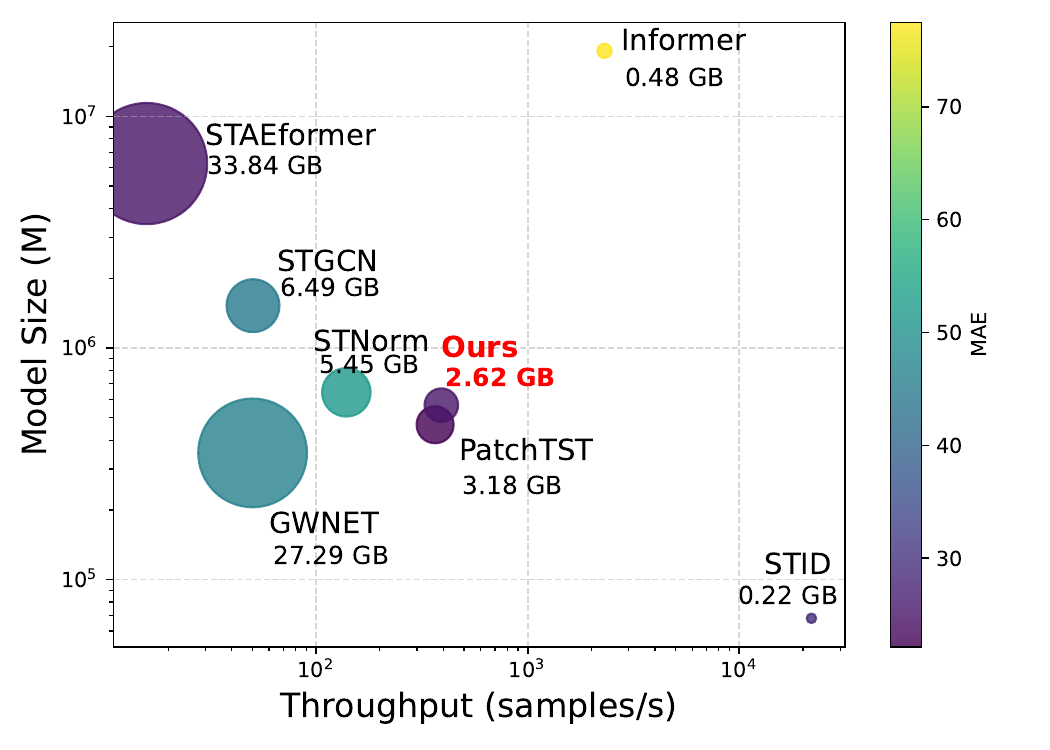}
        \caption{TaxiBJ}
        \label{fig:sub2}
    \end{subfigure}
    \begin{subfigure}[b]{0.32\textwidth}
        \centering
        \includegraphics[width=\textwidth]{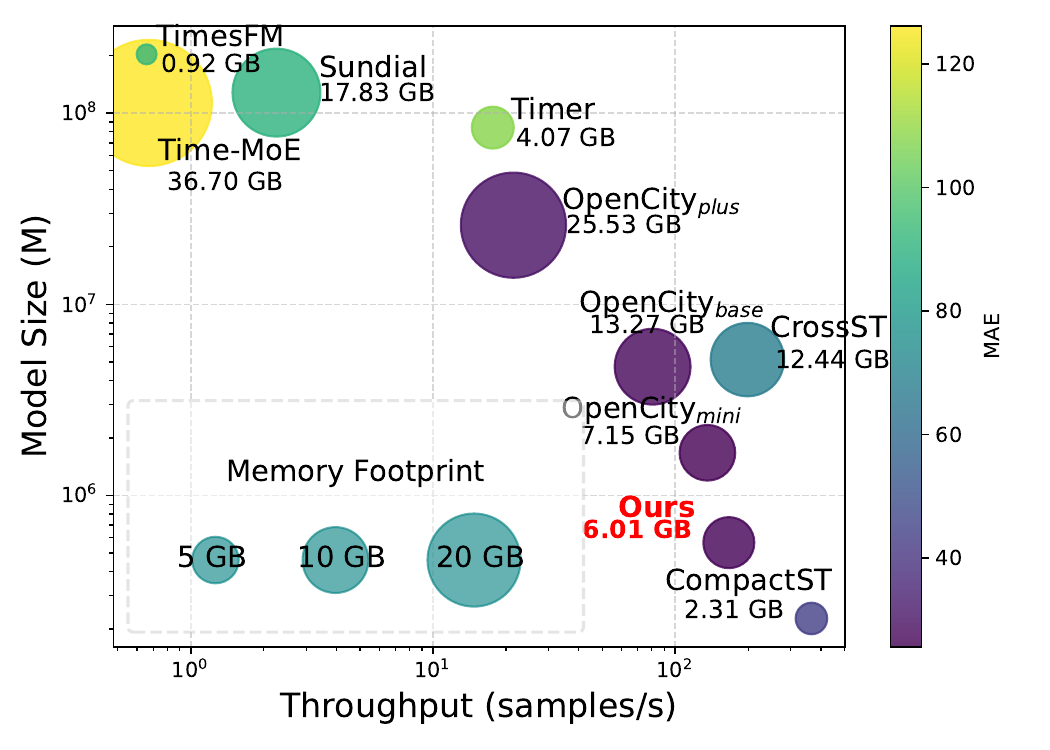}
        \includegraphics[width=\textwidth]{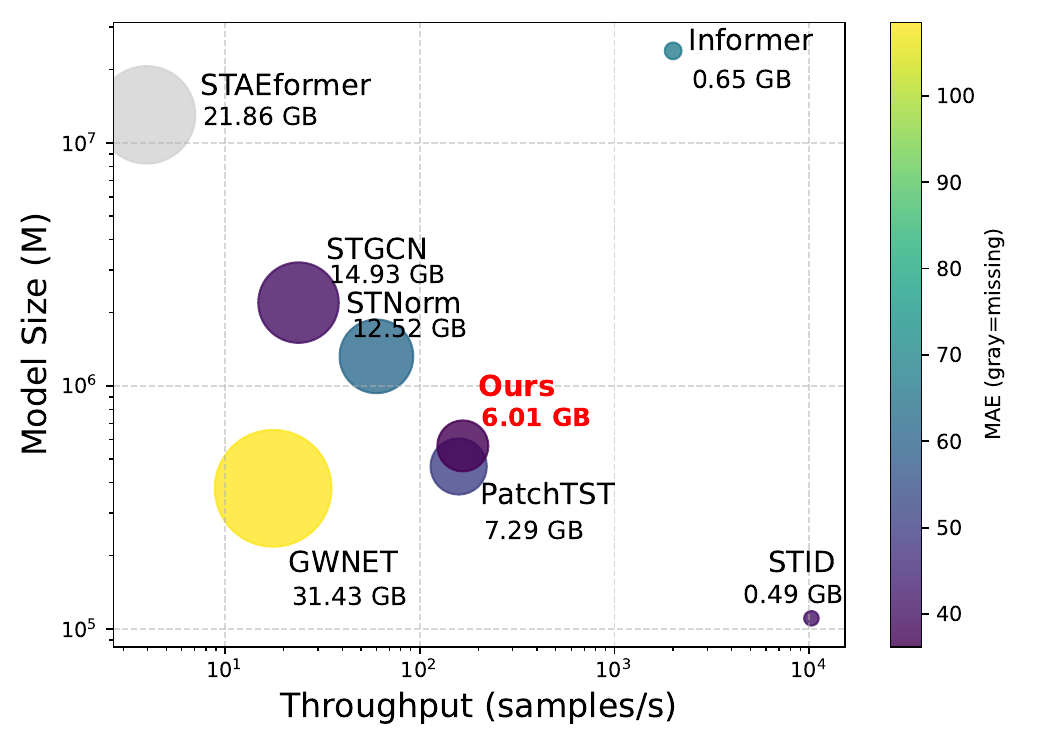}
        \caption{GBA}
        \label{fig:sub3}
    \end{subfigure}

    \caption{Peak memory, inference throughput and model parameter counts.}
    \label{fig:three_figures}
    \label{fig:param}
\end{figure*}

\subsection{Overall Results}
\subsubsection{Zero-shot Prediction}  
We evaluate the zero-shot performance of \nm\ under short-term (12-step) and long-term (64-step) forecasting settings, following \cite{unist, opencity}. Short-term forecasting covers time spans from hourly to half-day, while long-term forecasting spans from half-day to one day, with both ranges depending on the dataset sampling intervals. These settings support practical tasks such as short-term traffic monitoring and demand prediction for short-term horizons, as well as daily public transit scheduling and maintenance resource allocation for long-term horizons. As shown in Tables \ref{tab:zero_short} and \ref{tab:zero_long}, \nm\ achieves consistently competitive or superior performance across both horizons. In short-term prediction, \nm\ reduces predictive errors by up to 17.77\% compared to the second-best methods, with only a marginal 3.02\% increase in RMSE. Its advantages become more pronounced in long-term forecasting: on HZMetro inflow, \nm\ reduces error by 39.96\% in the short-term and by 60.18\% in the long-term compared to HA. Overall, time-series models underperform spatio-temporal approaches, highlighting the necessity of modeling spatial dependencies. Strong baselines such as CrossST and CompactST use shared pattern banks or global normalization to capture inconsistent dynamics within the homogeneous modeling paradigm. CrossST relies on learned patterns, whereas CompactST tends to oversimplify multi-domain forecasting. Both methods ignore periodic patterns, which reduces their accuracy in long-term forecasting. In contrast, \nm\ models multi-resolution spatio-temporal dynamics, multi-domain heterogeneity, and cross-dataset periodicity, enabling more effective zero-shot generalization.

\subsubsection{Few-shot Prediction} 
We conduct few-shot experiments on unseen datasets using 7 days of training data, sufficient to capture weekly traffic patterns. For each dataset, we finetune \nm\ by updating the predictor and the final encoder layer. Tables~\ref{tab:few12} and \ref{tab:few64} report short- and long-term results. \nm\ consistently improves performance even with minimal data. In short-term prediction, it surpasses the second-best method by 0.49\%–15.11\% on seen tasks and 1.14\%–30.98\% on unseen tasks. Long-term prediction shows similar gains, demonstrating that \nm\ captures periodic patterns effectively with limited finetuning. Baselines show mixed robustness. GNN-based methods, such as STGCN and GWNET, model spatial correlations but decline on long-term horizons due to lack of periodic modeling. Methods relying on learnable random features, such as STID and STAEformer, are sensitive to data sparsity. Overall, \nm\ maintains strong performance across datasets, benefiting from its modular, adaptive design that jointly models spatial, temporal, and periodic patterns, enabling robust knowledge transfer.

\subsection{Scalability and Data Efficiency}

\subsubsection{Scalability Analysis} 
We evaluate \nm’s scalability in peak GPU memory, inference throughput, and parameter count on three datasets: PEMS08 (170 nodes), Beijing Taxi (1,024 nodes), and GBA (2,352 nodes), using a batch size of 64. Baselines encountering out-of-memory (OOM) errors reduce batch size to fit GPU memory. Figure~\ref{fig:param} shows that \nm\ maintains competitive scalability while preserving robust zero- and few-shot performance. TimesFM uses less memory but is slow; CompactST, CrossST, Informer, and STID are memory-efficient but sometimes unstable; Timer, Sundial, Time-MoE, and STAEformer require large memory, reducing throughput as node counts increase. \nm\ lowers computational cost by modeling spatial dependencies via global summaries and employing a sparse MoE design, reducing multi-resolution modeling overhead and enabling fast, efficient inference.

\subsubsection{Data Efficiency Analysis} 
We evaluate \nm’s adaptability in data-scarce scenarios by performing few-shot prediction with varying training volumes (Figure~\ref{fig:ft}). Even with a single day of data, \nm\ shows notable improvement, with gains increasing as more data becomes available. This demonstrates that its modular design effectively transfers shared knowledge, isolates incompatible representations, and aligns coherent subspaces, enabling rapid capture of temporal, spatial, and periodic patterns with minimal supervision.
\begin{figure}[ht]
    \centering
    \includegraphics[width=\linewidth]{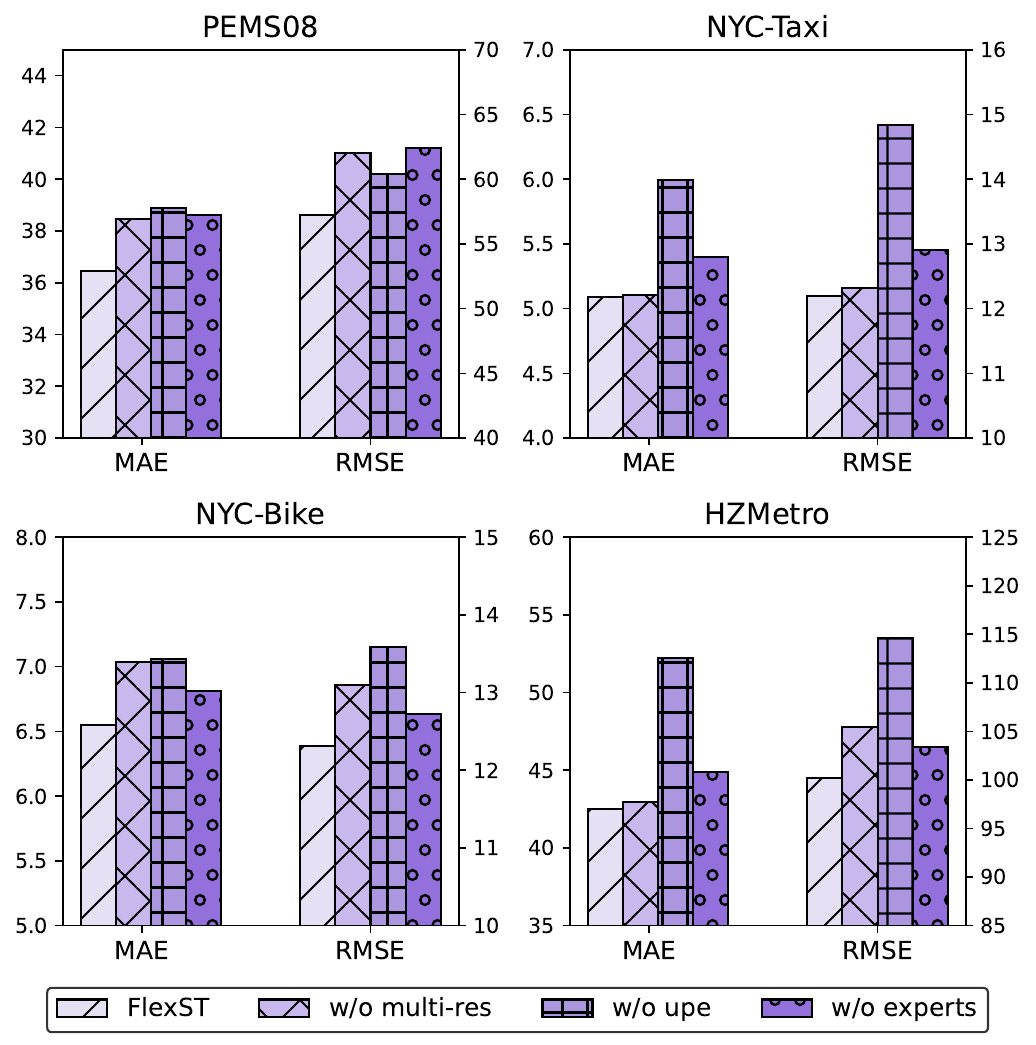}
    \caption{Ablation study.}
    \label{fig:ab}
\end{figure}
\subsection{Ablation Study}
We perform an ablation study of \nm\ on PEMS08, NYC-Taxi, NYC-Bike, and HZMetro in a zero-shot setting. Three variants are evaluated: (1) \emph{w/o multi-res}, disabling multi-resolution patching and graph generation by using a single temporal resolution and replacing $\mathcal{D}$ with the adjacency matrix $\mathbf{A}$; (2) \emph{w/o experts}, removing temporal and spatial domain-adaptive experts; (3) \emph{w/o upe}, excluding the unified periodic encoding. Results in Figure~\ref{fig:ab} show that all components contribute differently. Removing unified periodic encoding causes the largest performance drop, particularly in cross-year (NYC-Taxi) and cross-city (HZMetro) predictions. The domain-adaptive mixture-of-experts consistently improves performance by selecting suitable experts, aiding tasks with partially unseen characteristics, such as on NYC-Bike.. Multi-resolution modeling is especially important for NYC-Bike, indicating that capturing spatio-temporal patterns at multiple resolutions enhances generalization to novel scenarios.
\section{Conclusion}
In this paper, we present \nm, a modular and adaptive pre-training framework for robust zero-shot and few-shot forecasting of multi-domain traffic time series. \nm\ adopts a modular design and integrates three core components: (1) a multi-resolution spatio-temporal diffusion module that captures both short-term fluctuations and long-term trends while modeling cross-resolution spatio-temporal dependencies; (2) a domain-adaptive mixture-of-experts that dynamically routes inputs to specialized temporal and spatial experts, isolating incompatible patterns while enabling selective knowledge transfer; and (3) a unified periodic encoding strategy that injects resolution- and domain-aware periodic priors to harmonize cross-dataset periodic inconsistencies. Extensive experiments on 23 traffic datasets demonstrate that \nm\ achieves notable performance gains for both short-term and long-term predictions under zero-shot and few-shot settings, with competitive efficiency in terms of memory usage and inference throughput. These results highlight \nm’s ability to generalize across multi-domain traffic scenarios, offering a practical and effective solution for large-scale spatio-temporal forecasting in real-world deployments.

\section*{Acknowledgements}
This work was supported by the National Natural Science Foundation of China under Grant Nos. 62572417, 62421002, and 62602691, and by the CCF-DiDi GAIA Collaborative Research Funds.
\bibliographystyle{IEEEtran}
\bibliography{references}

\newpage

\section*{Appendix}
\begin{table}[t]
\centering
\caption{Summary of notations used in this paper.}
\label{tab:notation}
\resizebox{\linewidth}{!}{ \renewcommand{\arraystretch}{1.0}
\footnotesize 
\begin{tabularx}{\linewidth}{lX}
\toprule
\textbf{Notation} & \textbf{Description} \\
\midrule
$G=(\mathcal{V}, \mathbf{A})$ & Graph with spatial units $\mathcal{V}$ and adjacency matrix $\mathbf{A}$ representing spatial connections \\
$\mathbf{X}\in\mathbb{R}^{N\times T}$ & Single-domain traffic data with $T$ observations per spatial unit $n\in N$, characterized by task type $Q$, temporal resolution $R$, spatial resolution $D$ and spatial graph $G$ \\
$S$ & Time span of time series, determining its absolute duration \\
$\mathbf{ts}$ & timestamps associated with observations \\
$D, \mathbf{D}_{i}$ & Spatial resolution and spatial receptive field at resolution $i$ \\
$\mathcal{X}$ & Multi-domain traffic dataset\\
$\mathbf{P}_i^
\prime \in\mathbb{R}^{N\times M_i \times len_i}$ & Collection of patches at resolution $i$, containing $M_i$ patches, $len_i$ is the length of each patch \\
$\mathbf{P}^l_i\in\mathbb{R}^{N\times M_i\times d}$ & Patch representations at resolution $i$ for layer $l$ \\
$\mathbf{Z}^l\in\mathbb{R}^{N\times 1 \times d}$ & Collection of global summaries for nodes at layer $l$ \\
$\mathcal{E}^{l}_{Q}, \mathcal{E}^{l}_{R}, E^l_{share}$ & Task-gated, resolution-gated, and shared experts at layer $l$ \\
$\mathbf{H}_{pos}, \mathbf{H}_{Q}^l, \mathbf{H}_{R}^l, \mathbf{H}_{\mathbf{ts}}^l$ & Positional, task, resolution, and timestep encodings \\
\bottomrule
\end{tabularx}}
\end{table}

\section{Notations}
Table~\ref{tab:notation} summarizes the main notations used throughout the paper. 

\begin{figure}[!ht]
    \centering
    \begin{subfigure}{1\linewidth}
    \centering
    \includegraphics[width=0.49\linewidth]{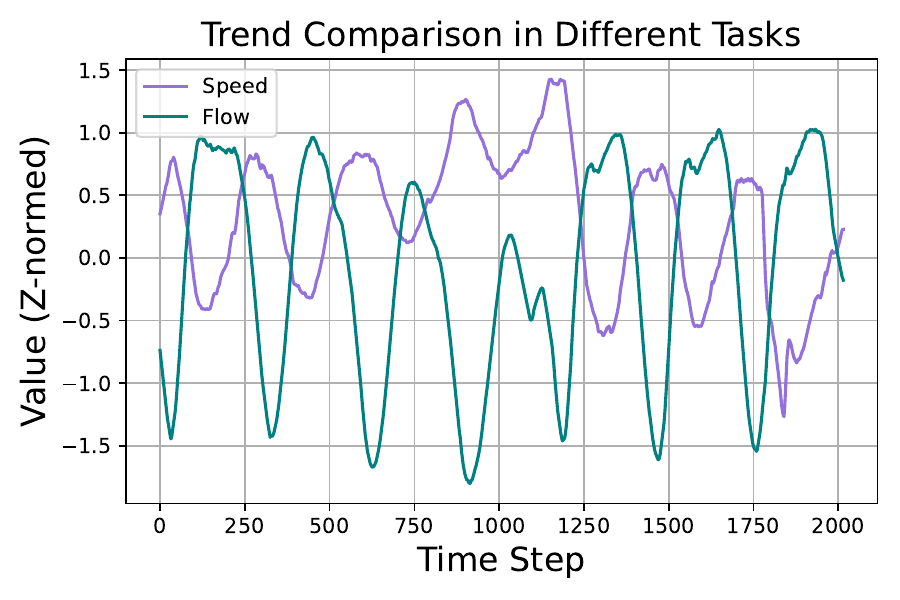}
    \includegraphics[width=0.49\linewidth]{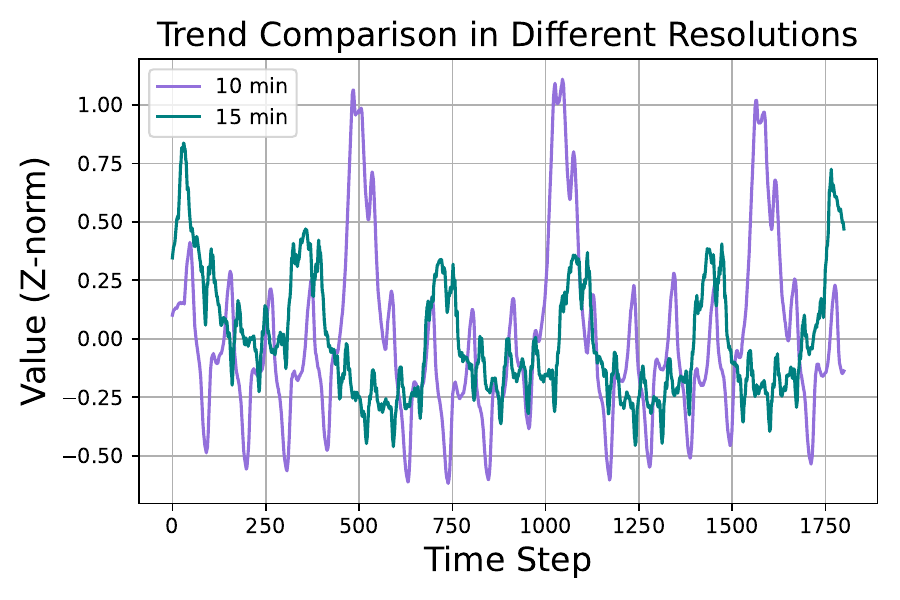}
    \caption{Temporal heterogeneity across tasks (PEMS08 flow vs. speed) and resolutions (Beijing Subway 15 min vs. 30 min). Values are Z-score normalized to reduce range differences.}\label{fig:temporal}
    \end{subfigure}
    \begin{subfigure}{1\linewidth}
    \includegraphics[width=0.49\linewidth]{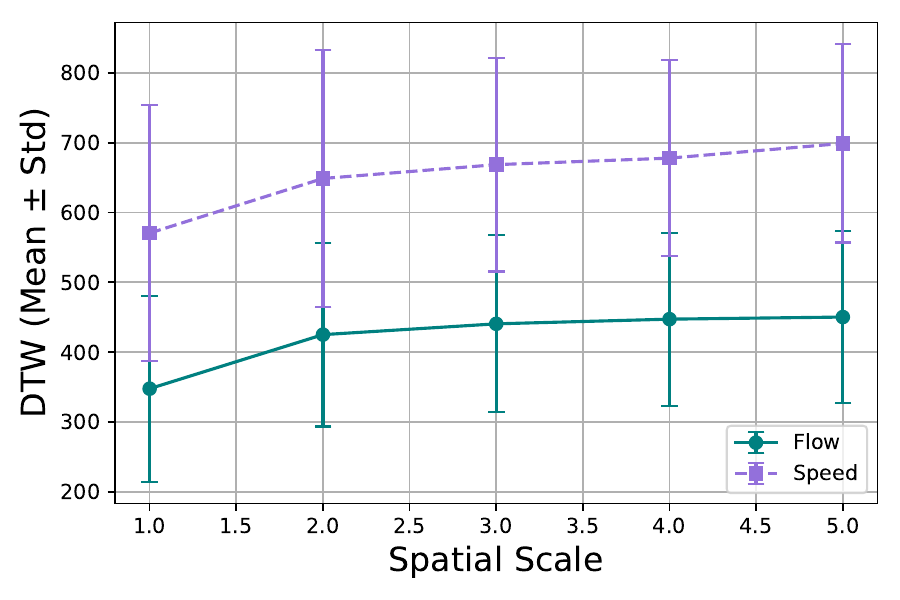}
    \includegraphics[width=0.49\linewidth]{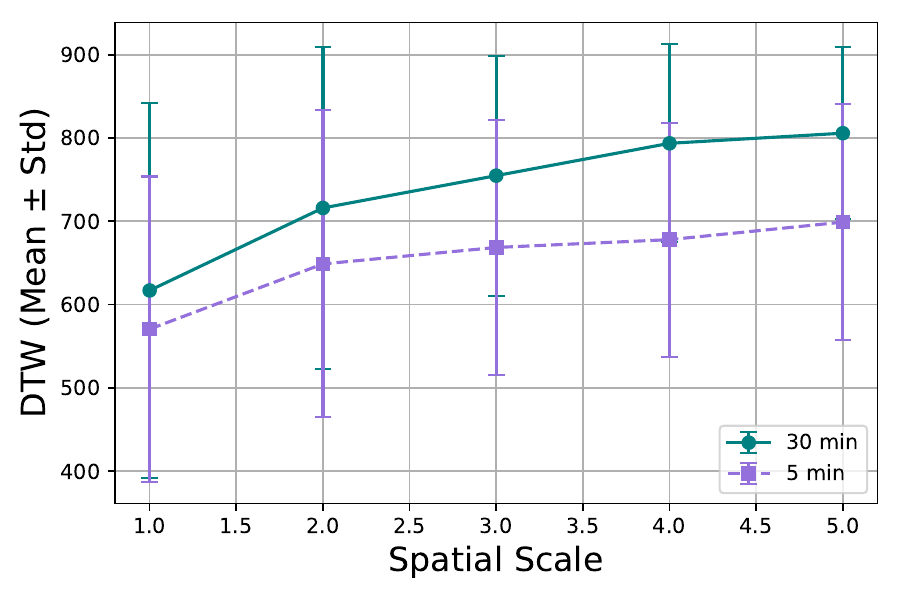}
    \caption{Spatial heterogeneity across tasks (PEMS08 flow
vs. PEMSBAY speed) and resolutions (30 min TrafficJN vs.
5 min PEMSBAY). }\label{fig:spatial}
    \end{subfigure}
    \caption{ Spatio-temporal dynamics observed across different forecasting tasks (left) and temporal resolutions (right).}
    \label{fig:motivation}
\end{figure}

\section{Data Analysis} \label{app:motivation}

To highlight the challenges of existing large-scale pre-trained spatio-temporal models, we conduct an empirical study focusing on two sources of heterogeneity: forecasting task and temporal resolution. We sample nodes in the spatial graph under different predictive variables and resolutions, and visualize the corresponding time series in Figure~\ref{fig:temporal}. For comparability, series are aligned by time step (index), reflecting a fixed-length input window. We also compute the mean and standard deviation of Dynamic Time Warping (DTW) distances between each node and its neighbors, measuring time series dissimilarity. Variation of these distances with hop count is shown in Figure~\ref{fig:spatial}, capturing how heterogeneity affects spatial dependencies. Results in Figure~\ref{fig:temporal} show that, although time series across tasks or resolutions share broadly similar trends and exhibit periodic patterns, they demonstrate heterogeneous temporal dynamics and inconsistent periodicity, with distinct pattern shapes, phase shifts, and amplitude variations across periodic cycles. Figure~\ref{fig:spatial} shows that DTW distances generally increase with hop count, but their growth rates differ across tasks and resolutions, revealing heterogeneous spatial dynamics. These observations motivate the design of a modular architecture that separates dataset-specific variations while aligning transferable knowledge.

\section{Complexity Analysis} \label{app:complexity}
For multi-resolution spatio-temporal diffusion, producing patch representations has complexity $\mathcal{O}(\sum_{i=1}^{K} N \cdot M_i \cdot \text{len}_i \cdot d) $, which simplifies to $\mathcal{O}(N \cdot T \cdot d)$ since $\sum_{i=1}^{K} M_i \cdot \text{len}_i \approx T$, where $M = \sum_{i=1}^{K} M_i$. Computing graph summaries adds $\mathcal{O}(N \cdot T \cdot d)$, while constructing propagation matrices for $\Delta$ predefined diffusion scales requires $\mathcal{O}(\Delta \cdot N^3)$, performed once before training. For an encoder with $L$ layers, per-layer costs are: spatial diffusion $\mathcal{O}(N^2 \cdot d)$ and its associated MLP projection $\mathcal{O}(N \cdot d^2)$, temporal self-attention $\mathcal{O}(\sum_{i=1}^{K} M_i^2 \cdot d)$, temporal decoupling experts $\mathcal{O}(N \cdot M \cdot (2k+1) \cdot d^2)$, top-$1$ matrix selection $\mathcal{O}(\Delta)$, and embedding lookups and projections of unified periodic encoding $\mathcal{O}(N \cdot M \cdot d^2)$. Multiplying by $L$ layers, the total encoder time complexity becomes $\mathcal{O}(L \cdot (N^2 \cdot d + N \cdot d^2 + \sum_{i=1}^{K} M_i^2 \cdot d + N \cdot M \cdot (2k+1) \cdot d^2 + N \cdot M \cdot d^2 + \Delta))$. Including patch embedding and graph summary computations, and considering bounded $k$ and small $\Delta$, the dominant total complexity simplifies to $
\mathcal{O}(N \cdot T \cdot d + L \cdot (N^2 \cdot d + \sum_i M_i^2 \cdot d + N \cdot M \cdot d^2))$. Among the quadratic terms, the $N^2$ component arises from spatial diffusion and can be reduced toward linear or near-linear complexity using techniques such as sparse graph representations or low-rank approximations of propagation matrices. The quadratic cost from multi-resolution patch interactions remains inherent, but it is typically manageable since $M_i \ll T$. Overall, the model remains efficient for large-scale multi-resolution spatio-temporal forecasting.

\section{Evaluations}
\subsection{Datasets}
This section provides a brief description of the datasets used for pre-training and downstream evaluation, which differ in spatial regions, temporal resolutions, tasks, and time spans.
\begin{itemize}

\item \textbf{PEMS03} \cite{Song2020STSGCN}:
Traffic flow data collected every 5 minutes from California, USA, spanning September 1 to November 30, 2018, with 358 sensor nodes.

\item \textbf{PEMS04} \cite{Song2020STSGCN}:
Traffic flow data from San Francisco, USA, sampled every 5 minutes between January 1 and February 28, 2018, consisting of 307 nodes.

\item \textbf{PEMS07} \cite{Song2020STSGCN}:
Traffic flow data collected every 5 minutes in California, USA, from May 1 to August 31, 2017, with 883 nodes.

\item \textbf{PEMS08} \cite{Song2020STSGCN}:
Traffic flow data from San Bernardino, USA, sampled every 5 minutes between July 1 and August 31, 2016, containing 170 nodes.

\item \textbf{SD} \cite{largest}:
Traffic flow data collected every 5 minutes in San Diego, USA, covering the year 2020, with 716 nodes.

\item \textbf{GBA} \cite{largest}:
Large-scale traffic flow data from the Bay Area, USA, sampled every 5 minutes throughout 2020, consisting of 2,352 nodes.

\item \textbf{NYC-Taxi} \cite{UrbanGPT}:
Taxi demand data from New York City, USA, sampled every 30 minutes, covering 2016 for pre-training and 2017–2021 for evaluation, with 263 regions.

\item \textbf{Beijing Taxi} \cite{unist}:
Taxi demand data collected every 30 minutes in Beijing, China, spanning multiple periods between 2013 and 2016, with 1,024 regions.

\item \textbf{CHI-Taxi} \cite{UrbanGPT}:
Taxi demand data from Chicago, USA, sampled every 30 minutes throughout 2021, consisting of 77 regions.

\item \textbf{NYC-Bike} \cite{UrbanGPT}:
Bike-sharing demand data from New York City, USA, sampled every 30 minutes between 2016 and 2021, with 540 stations.

\item \textbf{CHI-Bike} \cite{libcity}:
Bike-sharing data collected every 30 minutes in Chicago, USA, from July to September 2020, consisting of 270 stations.

\item \textbf{TrafficHZ} \cite{unist}:
Traffic speed data from Hangzhou, China, sampled every 30 minutes between March 5 and April 5, 2022, with 672 road segments.

\item \textbf{TrafficJN} \cite{unist}:
Traffic speed data collected every 30 minutes in Jinan, China, from March 5 to April 5, 2022, consisting of 576 nodes.

\item \textbf{TrafficSH} \cite{unist}:
Traffic speed data from Shanghai, China, sampled every 30 minutes between January 27 and February 27, 2022, with 896 nodes.

\item \textbf{TrafficZZ} \cite{unist}:
Traffic speed data collected every 30 minutes in Zhengzhou, China, between March and June 2012, consisting of 676 nodes.

\item \textbf{PEMSBAY} \cite{li2018diffusion}:
Traffic speed data from the Bay Area, USA, sampled every 5 minutes between January and May 2017, with 325 sensors.

\item \textbf{METR-LA} \cite{li2018diffusion}:
Traffic speed data collected every 5 minutes in Los Angeles, USA, from March to June 2012, consisting of 207 sensors.

\item \textbf{Beijing Subway (10 min)} \cite{subway_bejing}:
Metro flow data from Beijing, China, sampled every 10 minutes between February 29 and April 3, 2016, with 276 stations.

\item \textbf{Beijing Subway (15 min)} \cite{subway_bejing}:
Metro flow data from Beijing, China, sampled every 15 minutes over the same period, consisting of 276 stations.

\item \textbf{SHMetro} \cite{metro}:
Metro flow data collected every 15 minutes in Shanghai, China, from July to September 2016, with 288 stations.

\item \textbf{HZMetro} \cite{metro}:
Metro flow data from Hangzhou, China, sampled every 15 minutes between January 1 and January 25, 2019, consisting of 80 stations.

\item \textbf{DIDI-CD} \cite{didi}:
Traffic congestion level data collected every 10 minutes in Chengdu, China, from January to April 2018, with 524 regions.

\item \textbf{DIDI-SZ} \cite{didi}:
Traffic congestion level data from Shenzhen, China, sampled every 10 minutes over the same period, consisting of 627 regions.

\end{itemize}
Tables ~\ref{tab:pretain} and \ref{tab:eval} summarizes the key statistics of the datasets used in pre-training and downstream evaluations.
\begin{table*}[ht]
\renewcommand\arraystretch{1.0}
    \centering
    \caption{Statistical information of the pre-training datasets, which are aggregated based on the measured variables. For sampling rate: T=minute, H=hour. We provide detailed description of each dataset in supplementary material.}\label{tab:pretain}
    \scalebox{1.0}{
    \begin{tabular}{c c c c c c c}
        \toprule
        \multicolumn{1}{c}{\textbf{Dataset}} & \textbf{Traffic Flow} & \textbf{Taxi Demand} & \textbf{Traffic Speed} & 
        \textbf{Metro Flow} &
        \textbf{Congestion Level} & \textbf{Total}\\
        \midrule
        \multicolumn{1}{c}{\# of Regions} & 1,957 
        & 263 
        & 2,676 
        & 564
        & 524
        & 3,584  \\
        \multicolumn{1}{c}{\# of Records} & 71,834,110 
        & 9,240,768 
        & 28,015,352 
        & 5,621,616 
        & 9,054,720
        & 123,766,566\\
        \multicolumn{1}{c}{Sampling Rate} & 5T & 30T & 5T, 30T & 10T, 15T & 10T & 5T, 10T, 15T, 30T \\
        \multicolumn{1}{c}{Source} & \cite{Song2020STSGCN,largest} & \cite{UrbanGPT} & \cite{unist,li2018diffusion} & \cite{subway_bejing,metro} & \cite{didi} & - \\
        \bottomrule
    \end{tabular}
    }
\end{table*}

\begin{table*}[ht]
    \centering
    \caption{Statistical information of the downstream evaluation datasets. Notice that, the evaluation datasets differ from the pre-training datasets in terms of \emph{spatial regions, sampling rates, tasks, or years}.}\label{tab:eval}
    \resizebox{\textwidth}{!}{ \renewcommand{\arraystretch}{1.1}
    \setlength{\tabcolsep}{2pt}
    \begin{tabular}{>{\centering\arraybackslash}m{2cm} 
                >{\centering\arraybackslash}m{1.5cm} 
                >{\centering\arraybackslash}m{1.5cm} 
                >{\centering\arraybackslash}m{1.5cm} 
                >{\centering\arraybackslash}m{1.5cm} 
                >{\centering\arraybackslash}m{1.5cm} 
                >{\centering\arraybackslash}m{1.5cm} 
                >{\centering\arraybackslash}m{1.5cm} 
                >{\centering\arraybackslash}m{1.5cm} 
                >{\centering\arraybackslash}m{1.5cm} 
                >{\centering\arraybackslash}m{1.5cm} 
                >{\centering\arraybackslash}m{1.5cm} 
                >{\centering\arraybackslash}m{1.5cm}}
        \toprule
        \multicolumn{1}{c}{\textbf{Dataset}} & \textbf{PEMS04} & \textbf{PEMS08} & \textbf{GBA} & \textbf{NYC-Taxi} & \textbf{CHI-Taxi} & \textbf{Beijing Taxi} & \textbf{NYC-Bike} & \textbf{CHI-Bike} & \textbf{TrafficZZ} & \textbf{Beijing Subway} & \textbf{HZMetro} & \textbf{Didi-SZ}\\
        \midrule
        \multicolumn{1}{c}{Task} & Traffic Flow & Traffic Flow & Traffic Flow & Taxi Demand & Taxi Demand & Taxi Demand & Bicycle Sharing & Bicycle Sharing & Traffic Speed & Metro Flow & Metro Flow & Congestion Level \\
        \multicolumn{1}{c}{\# of Regions} & 307 & 170 & 2,352 & 263 & 77 & 1,024 & 540 & 270 & 676 & 276 & 80 & 627 \\
        \multicolumn{1}{c}{\# of Records} & 5,216,544 & 3,035,520 &  124,637,184 & 46,102,848 & 2,698,080 & 43,745,280 & 113,633,280 & 2,384,640 & 948,428 & 993,600 & 292,000 & 10,834,560 \\
        \multicolumn{1}{c}{Sampling Rate} & 5T & 5T & 5T & 30T & 30T & 30T & 30T & 30T & 30T & 15T & 15T & 10T \\
        \multicolumn{1}{c}{Source} & \cite{Song2020STSGCN} & \cite{Song2020STSGCN} & \cite{largest} & \cite{UrbanGPT} & \cite{UrbanGPT} & \cite{unist} & \cite{UrbanGPT} & \cite{libcity} & \cite{unist} & \cite{li2018diffusion} & \cite{metro} & \cite{didi}\\
        \bottomrule
    \end{tabular}}
\end{table*}

\subsection{Hyperparameter Analysis}

We performed an across-dataset trend analysis on 12 target datasets, using per-dataset relative change vs. default and then averaging across datasets. With fixed total experts, we vary top-K active experts. For temporal top-K (K=1/2/3), K=2 is near-neutral (+0.18\% MAE, +0.31\% RMSE), while K=3 is clearly worse (+0.95\% MAE, +1.17\% RMSE). For spatial top-K, both K=2 and K=3 degrade performance relative to K=1 (+1.69\%/+ 2.92\% and +0.97\%/+2.10\% in MAE/RMSE). In the joint temporal+spatial setting, K=2 performs best overall (-0.19\% MAE, -0.26\% RMSE), while K=3 again degrades (+0.67\% MAE, +0.55\% RMSE). Increasing diffusion upper bound from 3 to 6 also worsens performance (+3.06\% MAE, +3.81 \% RMSE). These trends indicate that moderate routing sparsity is more robust, whereas aggressive settings are not consistently beneficial. Heat diffusion matrices are precomputed once and reused, so the cost is offline preprocessing, not per-iteration training/inference runtime. Measured preprocessing time: GPU 0.01s (1k nodes), 0.07s (2k), 1.10s (5k), 9.05s (10k); CPU 0.39s, 3.37s, 54.53s, 396.09s.

\subsection{Routing Analysis}

We analyze top-K routing under K = 1/2/3 over 24 task-scale conditions (6 tasks $times$ 4 scales). In both Temporal and Spatial decoupling, task logits consistently separate Traffic Flow from other task families. In temporal-scale routing, layer-1 keeps the same top-1 expert across tasks/scales, while layer-0/2 vary with scale. In spatial-scale routing, the layer-2 top1-top2 logit gap increases from 5-min to 30-min within each K (K=1: 0.01 $\rightarrow$ 0.05; K=2: 0.16 $\rightarrow$ 0.96; K=3: 0.20 $\rightarrow$ 0.98), indicating structured scale-conditioned routing. Top-K selected-expert coverage is temporal-task 4/8, 7/8, 8/8; temporal-scale 3/8, 4/8, 5/8; spatial 4/13, 8/13, 9/13 (K=1/2/3). We acknowledge some experts are rarely selected, which is expected in sparse overcomplete MoE and does not indicate routing collapse: routing remains structured by task family and scale, and only top-K experts run per sample, so low-usage experts can be pruned at deployment if needed.

\end{document}